\documentclass[sigconf]{acmart}

\usepackage{graphicx}
\usepackage{dsfont}
\usepackage{bm}
\usepackage{booktabs}

\usepackage{amsmath,amsthm,amssymb}
\usepackage{mathrsfs}
\usepackage{arydshln} % 加载虚线支持包
\usepackage{subcaption}
\usepackage{wrapfig}
\usepackage{multirow}
\usepackage{enumitem} % 控制itemize行宽
\usepackage[table]{xcolor} % 用于表格颜色
\definecolor{rowgray}{gray}{0.9}

\usepackage{algorithmic}
\usepackage{algorithm}

\usepackage{subfiles}

\newtheorem{lemma}{Lemma}

\newtheorem{definition}{Definition}

\copyrightyear{2026}
\acmYear{2026}
\setcopyright{cc}
\setcctype{by}
\acmConference[KDD 2026] {Proceedings of the 32nd ACM SIGKDD Conference on Knowledge Discovery and Data Mining V.2}{August 9--13, 2026}{Jeju Island, Republic of Korea.}
\acmBooktitle{Proceedings of the 32nd ACM SIGKDD Conference on Knowledge Discovery and Data Mining V.2 (KDD 2026), August 9--13, 2026, Jeju Island, Republic of Korea}
\acmISBN{979-8-4007-2259-2/2026/08}
\acmDOI{10.1145/3770855.3817653}
\begin{document}

\title{STM3: Mixture of Multiscale Mamba for Long-Term Spatio-Temporal Time-Series Prediction}

\settopmatter{authorsperrow=4}

\author{Haolong Chen}
\affiliation{%
  \institution{Shenzhen International\\Center for Industrial and\\Applied Mathematics}
  % \city{Shenzhen}
  % \state{Guangdong}
  % \country{}
  \city{Shenzhen}
  % \state{Guangdong}
  \country{China}
}
\affiliation{%
  \institution{Shenzhen Research\\Institute of Big Data}
  \city{Shenzhen}
  % \state{Guangdong}
  \country{China}
}
\affiliation{%
  \institution{The Chinese University\\of Hong Kong, Shenzhen}
  \city{Shenzhen}
  % \state{Guangdong}
  \country{China}
}
\email{haolongchen1@link.cuhk.edu.cn}

\author{Liang Zhang}
\authornote{Co-corresponding authors.}
\affiliation{%
  \institution{Shenzhen Campus of\\Sun Yat-sen University}
  \city{Shenzhen}
  % \state{Guangdong}
  \country{China}
}
\email{zhangliang27@mail.sysu.edu.cn}

\author{Zhengyuan Xin}
\affiliation{%
  \institution{Shenzhen Research\\Institute of Big Data}
  \city{Shenzhen}
  % \state{Guangdong}
  \country{China}
}
\affiliation{%
  \institution{The Chinese University\\of Hong Kong, Shenzhen}
  \city{Shenzhen}
  % \state{Guangdong}
  \country{China}
}
\email{zhengyuanxin@link.cuhk.edu.cn}

\author{Guangxu Zhu}
\authornotemark[1]
\affiliation{%
  \institution{Shenzhen International\\Center for Industrial and\\Applied Mathematics}
  \city{Shenzhen}
  % \state{Guangdong}
  \country{China}
}
\affiliation{%
  \institution{Shenzhen Research\\Institute of Big Data}
  \city{Shenzhen}
  % \state{Guangdong}
  \country{China}
}
\affiliation{%
  \institution{The Chinese University\\of Hong Kong, Shenzhen}
  \city{Shenzhen}
  % \state{Guangdong}
  \country{China}
}
\affiliation{%
  \institution{Shenzhen Loop Area Institute}
  \city{Shenzhen}
  % \state{Guangdong}
  \country{China}
}
\email{gxzhu@sribd.cn}

\renewcommand{\shortauthors}{Haolong Chen, Liang Zhang, Zhengyuan Xin, and Guangxu Zhu.}


\begin{abstract}
Recently, spatio-temporal time-series prediction has developed rapidly, yet existing deep learning methods struggle with learning complex long-term spatio-temporal dependencies efficiently. The long-term spatio-temporal dependency learning brings two new challenges: 1) The long-term temporal sequence naturally includes multiscale information, which is hard to extract efficiently; 2) The multiscale temporal information from different nodes is highly correlated and hard to model. To address these challenges, we propose \textit{\textbf{S}patio-\textbf{T}emporal \textbf{M}ixture of \textbf{M}ultiscale \textbf{M}amba} (STM3). STM3 integrates a Multiscale Mamba architecture within a novel Disentangled Mixture-of-Experts (DMoE) framework to capture diverse multiscale information efficiently, while utilizing an adaptive graph causal network to model complex spatial dependencies. To ensure robust representation learning, we introduce a stable routing strategy and a causal contrastive learning strategy, which work in tandem with hierarchical information aggregation to guarantee scale distinguishability. We theoretically prove that STM3 achieves superior routing smoothness and guarantees pattern disentanglement for each expert. 
Extensive experiments on 10 real-world benchmarks across domains demonstrate STM3's superior performance, achieving state-of-the-art results in long-term spatio-temporal time-series prediction. Notably, on the PEMSD8 dataset, it achieves significant improvements, surpassing the second-best model by \textbf{7.1\%} in MAE, \textbf{8.5\%} in RMSE, and \textbf{15.9\%} in MAPE. Code is available at \url{https://github.com/IfReasonable/STM3_KDD26}.
\end{abstract}


% \keywords{Do, Not, Use, This, Code, Put, the, Correct, Terms, for,




\maketitle

\section{Introduction}\label{sec:introduction}

Spatio-temporal time-series prediction is fundamental to applications like traffic forecasting and environmental monitoring~\cite{tang2025time, cini2025graph, chen2026overview}. Among these tasks, long-term prediction stands out as a critical yet challenging one, as it is essential for capturing the far-reaching propagation of spatial patterns over time. Numerous spatio-temporal and multivariate time-series neural networks with various types have been designed to address this task~\cite{zhao2019t, guo2019attention, liu2023spatio, yu2017spatio, li2018diffusion, song2020spatial}.

Recently, Mamba~\cite{gumamba} was introduced as a novel State Space Model (SSM)-based architecture, capable of efficiently modeling long-term dependencies through time-varying parameters. While preliminary attempts have applied Mamba to spatio-temporal data~\cite{li2024stg, choi2024spot, He_Ji_Lei_2025}, these approaches typically adopt an isolated perspective to model temporal and spatial dependencies, rather than viewing the information entanglement between these two dimensions through a unified lens. This limitation results in a significant decline in modeling accuracy, particularly when handling long-term sequences. For instance, STG-Mamba~\cite{li2024stg} incorporates local spatial adjacency with Mamba's ability to model temporal dependencies. Similarly, DST-Mamba~\cite{He_Ji_Lei_2025} performs multi-scale feature extraction along the temporal dimension while utilizing a bidirectional Mamba to capture spatial correlations. However, these methods largely overlook the intricate cross-dimensional interplay, insufficiently modeling the information flow mechanisms during feature fusion across complex long-term spatio-temporal patterns.

Modeling long-term spatio-temporal dependencies introduces two specific challenges. \textbf{1) Long-term sequences inherently contain multiscale information.} For instance, minute-level traffic data captures immediate patterns, whereas hour-level data reflects daily periodicities like rush hours, with these diverse dynamics coexisting within a single long-term sequence. While recent works such as DST-Mamba~\cite{He_Ji_Lei_2025} and TimeMixer++~\cite{wang2025timemixer++} attempt to extract complex multiscale information via time-series decomposition, their aggregation mechanisms remain coarse, often relying on naive summation or concatenation, which inevitably leads to the re-entanglement of multiscale information. Furthermore, approaches such as MixMamba~\cite{alkilane2024mixmamba} and SST~\cite{xu2025sst} utilize Mixture of Experts (MoE) to handle diverse temporal dynamics but perform routing at the time slot level. Consequently, slots from the same spatial node are fragmented across different experts. This instability negates the intended benefits of the architecture, leading to mode collapse where the experts fail to specialize and instead tend towards identical representations. \textbf{2) The multiscale temporal information across spatial nodes is highly correlated yet heterogeneous.} For example, traffic flow patterns differ significantly between highways and central business districts. It is hard to capture such diverse dynamics across different spatial locations. The challenge lies in effectively modeling these complex interactions and handling the distributional shifts between spatial nodes, which have not been fully explored. Although approaches like DGraFormer~\cite{yan2025dgraformer} utilize adaptive graph learning to model inter-node correlations, they capture local spatial dependencies and heterogeneity in a limited manner. Crucially, they fail to model the fundamental origin of spatial correlations—namely, that the similarity between spatial nodes is intrinsically driven by the similarity of their temporal patterns.

To address these challenges, we propose \textit{\textbf{S}patio-\textbf{T}emporal \textbf{M}ixture of \textbf{M}ultiscale \textbf{M}amba} (\textbf{STM3}), which systematically models long-term spatio-temporal dependencies by integrating three key innovations: 
\textbf{First, to address the multiscale temporal challenge,} we design \textbf{Multiscale Mamba}. Unlike existing multiscale approaches burdened by substantial computational overhead~\cite{guo2019attention, shabaniscaleformer} or limited by overly coarse aggregation mechanisms~\cite{He_Ji_Lei_2025, wang2025timemixer++}, our design efficiently reuses channels within a single Mamba block, augmented with learnable scale-specific biases. This allows the model to capture multiscale information while maintaining the computational advantages of Mamba.
\textbf{Second, to model intricate spatio-temporal dependencies,} we introduce \textbf{Adaptive Graph Causal Convolution Network (AGCCN)}. This module learns an adaptive graph structure shared across temporal scales and employs causal attention to regulate information flow. This ensures that each node aggregates features from relevant neighbors at the same or coarser scales while avoiding interference from finer scales, thereby preserving scale distinguishability. 
\textbf{Third, to resolve the heterogeneity challenge across spatial nodes,} we proposed a novel \textbf{Disentangled Mixture-of-Experts (DMoE)} mechanism. Since a single module may not generalize well to diverse spatial patterns (e.g., different between highway and urban traffic), we employ multiple expert models to partition and handle these intricate data distributions. Distinct from previous MoE approaches~\cite{shi2024time, xu2025sst, alkilane2024mixmamba}, we leverage static node embeddings rather than dynamic inputs for gating, resulting in significantly improved routing smoothness. Furthermore, we introduce a novel \textbf{causal contrastive learning} method that explicitly considers directional similarity, guaranteeing the disentanglement of patterns for each expert.

To summarize, our contributions lie in three aspects: 
\begin{itemize}[itemsep=0em, topsep=0em, leftmargin=1em] % 行宽，与上文的距离，与左端的距离
    \item We uncover the nature of spatio-temporal entanglement, revealing that inter-node spatial correlations are intrinsically driven by the similarity of their multiscale temporal patterns. Based on this, we introduce a method for multiscale feature aggregation and model decoupling, facilitating accurate long-term prediction.
    \item We introduce \textbf{STM3}, a comprehensive solution to spatio-temporal entanglement featuring a \textbf{tripartite innovation} on \textbf{temporal modeling} (Multiscale Mamba), \textbf{causal scale-wise spatio-temporal fusion} (AGCCN), and \textbf{expert disentanglement} (DMoE). Crucially, these components synergistically advance the state-of-the-art across these three critical dimensions.
    \item Comprehensive experiments on 10 benchmarks demonstrate that STM3 achieves SOTA performance.
\end{itemize}

\section{Related Work}
\paragraph{Spatio-Temporal Time-Series Prediction.} 
% Recent years have witnessed remarkable progress in spatio-temporal time-series prediction, driven by advances in deep neural networks specifically designed to capture complex spatial and temporal dependencies.
Deep neural networks have recently significantly advanced spatio-temporal prediction by capturing complex dependencies.
For temporal pattern modeling, researchers have developed models based on diverse architectures~\cite{zhao2019t, yu2017spatio, guo2019attention, liu2023spatio}. The recent emergence of Mamba-based approaches has introduced promising new directions in spatial correlation modeling~\cite{li2024stg, li2024dyg, yuan2025dg, choi2024spot}. Grounded in the theoretical foundations of SSMs, these methods demonstrate exceptional efficiency in processing sequential data. On the spatial modeling front, GNNs have become prevalent in capturing inter-region dependencies via graph-structured message passing~\cite{yu2017spatio, li2018diffusion, song2020spatial, He_Ji_Lei_2025, yan2025dgraformer}. Previous works such as DST-Mamba~\cite{He_Ji_Lei_2025} employed predefined graph structures for spatial relationships, or introduced adaptive graph learning methods—including DGraFormer~\cite{yan2025dgraformer}—which adaptively infer graph structures from data. In this work, we extend Mamba~\cite{gumamba} to construct a Multiscale Mamba architecture for spatio-temporal modeling. Our proposed approach also integrates adaptive graph causal convolution and a special MoE architecture, jointly capturing multiscale temporal dynamics and spatial dependencies.

More related work is summarized in the Appendix~\ref{appendix:more_related_work}.

\section{Problem Definition}

\paragraph{Spatio-Temporal Data.} Spatio-temporal data refer to data points collected from the real world within a specific region and time period. We represent it as ${X} = [x_{0}, x_{1}, \dots, x_{t}, \dots]$, where $x_{t} \in \mathbb{R}^{N \times C}$ denotes the recorded $C$ features of $N$ nodes at time step $t$.

% \paragraph{Spatial Graph.} We consider a spatial graph $\mathcal{G} = (\mathcal{V}, \mathcal{E}, \mathcal{A})$ to capture spatial relationships among nodes, where $\mathcal{V}$ is the set of $N$ nodes, $\mathcal{E}$ represents the set of edges, and $\mathcal{A}$ denotes the adjacency matrix of the graph, capturing proximity between nodes.

\paragraph{Spatio-Temporal Time-Series Prediction.} Spatio-temporal time-series prediction involves forecasting future spatio-temporal series based on historical observations. Given the historical data $X \in \mathbb{R}^{T \times N \times C}$, the goal of spatio-temporal time-series prediction is to estimate future data $Y \in \mathbb{R}^{\tau \times N \times C}$, where $T$ denotes the number of historical input timesteps and $\tau$ represents the prediction horizon.

\section{Methodology}\label{sec:methodology}
\begin{figure*}[t]
    \centering
    \includegraphics[width=\textwidth]{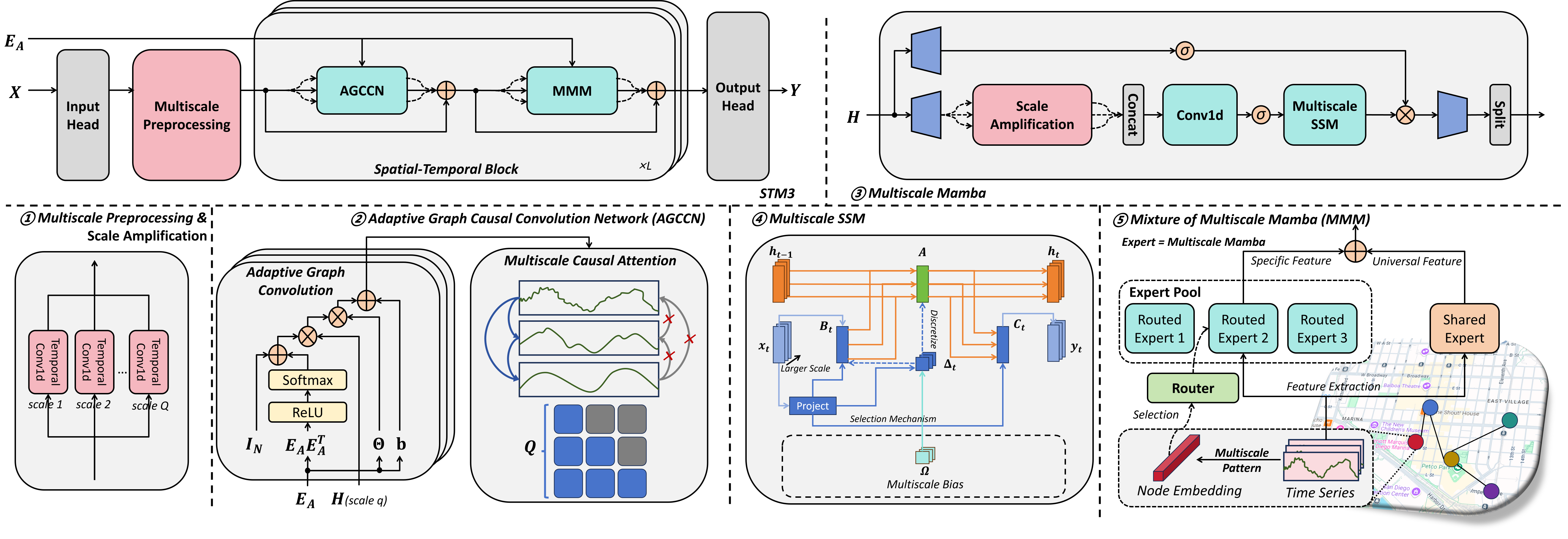}
    \caption{Main structure of STM3. 
    The spatio-temporal time series is first embedded and preprocessed into raw multiscale features, followed by $L$ stacked Backbone Layers. Each layer first utilizes an Adaptive Graph Causal Convolution Network (AGCCN) to process cross-scale and cross-node information propagation, and then employs a Disentangled Mixture of Multiscale Mamba (DMMM) to disentangle the multiscale temporal patterns of nodes and experts.}
    \vspace{-10pt}
    \label{fig:main}
\end{figure*}

We now introduce STM3, a disentangled multiscale mamba-based model for long-term spatio-temporal time-series prediction, as shown in Figure~\ref{fig:main}. It comprises three components: (1) Multiscale Preprocessing, (2) Adaptive Graph Causal Convolution Network, and (3) Disentangled Mixture of Multiscale Mamba.

\subsection{Multiscale Preprocessing}\label{multiscale_feature_Extraction}

The Multiscale Preprocessing step provides a basic multiscale representation of data. First, a linear layer as the input head projects the input $X \in \mathbb{R}^{T \times N \times C}$ into raw features $H \in \mathbb{R}^{T \times N \times d}$. Then, Multiscale Preprocessing module decomposes $H$ by multiple 1-d temporal convolution layers with varying kernel sizes: 
\begin{equation}
H^{(q)}_{\text{ms}} = \text{Conv1d}(H, s_0^{(q)}), \quad \forall q \in [1,Q],
\end{equation}
where $H^{(q)}_{\text{ms}} \in \mathbb{R}^{T \times N \times d}$ denotes the feature representation at the initial scale $s_0^{(q)} \in \mathbb{Z}^{+}$. We define scales such that $s_0^{(i)} \le s_0^{(j)}$ when $i \le j$, arranged from finer to coarser. $\text{Conv1d}(\cdot, s)$ indicates a 1-d temporal convolution with kernel size $s$. Finally, we stack all the features after convolution as the final output $H_{\text{ms}} \in \mathbb{R}^{T \times N \times d \times Q}$.

\subsection{Adaptive Graph Causal Convolution Network}\label{sec:AGCCN}
Adaptive Graph Causal Convolution Network (AGCCN) learns the complex correlation between the multiscale temporal information and the spatial information. There are two challenges: \textbf{1) The high computational complexity.} If we connect all nodes among all scales, the complexity is at least $\mathcal{O}(N^2Q^2Td)$, which is unacceptable when the problem becomes large enough. \textbf{2) The unknown hidden spatial relationship.} It is hard to obtain it from the real graph, such as the distance graph. Thus, we propose an adaptive causal convolution network with a learnable graph shared across all scales. Spatial information is first aggregated based on this graph, followed by causal attention for multiscale information aggregation. The complexity is then reduced to $\mathcal{O}(NQ^2Td)$.

We maintain a learnable node embedding dictionary $E_A \in \mathbb{R}^{N \times d_e}$ for all nodes, where each row of $E_A$ represents the embedding of a node and $d_e$ denotes the dimension of node embedding. Then, we can yield the normalized Laplacian directly:
\begin{equation}
    D^{-1/2}{A}D^{-1/2} = \text{softmax}(\text{ReLU}(E_AE_A^\top))
\end{equation}
where $A \in \mathbb{R}^{N\times N}$ is the adjacent matrix, and $D$ is the degree matrix. This allows joint graph structure learning with prediction objectives, overcoming domain-specific heuristic limitations.

Let $H_{\text{in}} \in \mathbb{R}^{N \times d \times Q}$ where $Q$ denotes the number of temporal scales denote the input traffic of a specific time step $t$, here we omit the subscript $t$ for brevity, and the $1^{st}$ order Chebyshev polynomial expansion approximated GCN operation follows:
\begin{equation}
\begin{aligned}
H_{\text{G}}^{(q)} &= (I_N + D^{-1/2} A D^{-1/2}) H_{\text{in}}^{(q)} \Theta + b, \\
\Theta &= E_g W_g, \quad b = E_g b_g, \quad \forall q \in [1, Q],
\end{aligned}
\end{equation}
where $I_N \in \mathbb{R}^{N \times N}$ is identity matrix, $\Theta \in \mathbb{R}^{N \times d\times d}$, $b \in \mathbb{R}^{N \times d}$ are learnable parameters, and $H_{\text{G}} \in \mathbb{R}^{N \times d \times Q}$ is the output of the adaptive graph convolution. For better optimization of the huge parameter matrices $\Theta$ and $b$, a low-rank adaptation approach is introduced as $\Theta=E_g W_g$, $b=E_gb_g$, where $d_{low} \ll N$, $E_g \in \mathbb{R}^{N\times d_{low}}$, $W_g \in \mathbb{R}^{d_{low}\times d\times d}$, $b_g \in \mathbb{R}^{d_{low}\times d}$ are learnable parameters.

After aggregating spatial information, $H_{\text{GCN}}$ is processed by causal attention to aggregate cross-scale information. The mechanism ensures hierarchical information aggregation, i.e., fine-grained features selectively retrieve coarser-grained information while preventing redundant reverse processes, thus avoiding information entanglement. The causal attention mechanism is as follows:
\begin{equation}
\begin{aligned}
\mathcal{Q} = \text{Linear}_\mathcal{Q}(H_{\text{G}}^\top), 
\mathcal{K} = \text{Linear}_\mathcal{K}(H_{\text{G}}^\top), 
\mathcal{V} = \text{Linear}_\mathcal{V}(H_{\text{G}}^\top), 
\end{aligned}
\end{equation}
\begin{equation}
\begin{aligned}
    H_{\text{atten}} &= \text{softmax}\left(\frac{\mathcal{Q} \mathcal{K}^\top}{\sqrt{d}} + \mathcal{M}\right)V, \quad \text{where} \\
    \mathcal{M}^{(i,j)} &= \begin{cases} 
    0 & \text{if } i \leq j \\
    -\infty & \text{otherwise}
    \end{cases} ,
\end{aligned}
\end{equation}
\begin{equation}
    H_{\text{out}} = \text{LayerNorm}(\text{Linear}_{\text{out}} (H_{\text{atten}})^\top + H_{\text{G}}),
\end{equation}
where $\text{Linear}_\mathcal{Q}$, $\text{Linear}_\mathcal{K}$, $\text{Linear}_\mathcal{V}$ and $\text{Linear}_{\text{out}}$ project the feature into the corresponding space. The upper-triangle causal attention mask $\mathcal{M} \in \mathbb{R}^{Q \times Q}$ constrains the rule ``only coarse-grained information can flow towards fine-grained information''. Finally, the feature after attention is projected to the output space, added with the residual input $H_{\text{GCN}}$, and normalized as the output feature $H_{\text{out}}$.

\subsection{Multiscale Mamba}\label{sec:multiscale_mamba}

Although the standard Mamba architecture effectively models long-term dependencies, it exhibits inherent limitations in simultaneously capturing multiscale temporal patterns. To address this, we propose a novel Multiscale Mamba module, detailed in Algorithm~\ref{alg:multiscale_mamba}. Let $H_{\text{in}}\in \mathbb{R}^{T \times d \times Q}$ denote the input feature under a specific spatial node $n$. Here we omit the subscript $n$ for brevity. After input projection, we can get the Mamba's latent feature $h_{\text{ms}} = \text{Linear}(H_{\text{in}})$.

\paragraph{Scale Amplification.} 
In the previous multiscale preprocessing module, a single input sequence is transformed into multiple sequences at $Q$ different temporal scales. Modeling long-term dependencies typically requires a large number of scale levels, which can incur high computational costs. To address this, we introduce a scale amplification mechanism that enables Mamba to capture additional temporal scales with low overhead. Structurally similar to multiscale preprocessing, it applies 1D temporal convolutions for scale amplification as follows:
% In the previous multiscale preprocessing module, we transformed a single input sequence into multiple sequences at different temporal scales, providing up to $Q$ distinct scales of temporal information. For modeling long-term temporal sequences, a large number of scale levels is typically required, which would lead to prohibitive computational complexity. To address this challenge, we introduce a scale amplification mechanism that enables the Mamba architecture to capture additional temporal scales. This mechanism is structurally similar to the multiscale preprocessing, employing 1-dimensional temporal convolution layers to perform scale amplification as follows:
\begin{equation}
    h^{(q)} = \text{Conv1d}(h_{\text{ms}}^{(q)}, s^{(q)}), \quad \forall q \in [1,Q],
\end{equation}
where $h_{\text{ms}}^{(q)}\in \mathbb{R}^{T \times d_{\text{inner}}}$ and $h^{(q)} \in \mathbb{R}^{T \times d_{\text{inner}}}$ denote the input and output feature sequences at scale $q$. We then stack the outputs to obtain $h \in \mathbb{R}^{T \times d_{\text{inner}} \times Q}$, with symbols consistent with Section~\ref{multiscale_feature_Extraction}. 
% Through scale amplification, the maximum scale expands to $s^{(Q)}_0 [s^{(Q)}]^L$, where $L$ represents the layer index in the backbone architecture where the Multiscale Mamba module is deployed, thereby establishing an exponentially growing temporal receptive field.
Through scale amplification, the maximum scale expands to $s^{(Q)}_0 [s^{(Q)}]^L$, where $L$ denotes the layer index of the backbone where the Multiscale Mamba module is deployed, enabling an exponentially growing temporal receptive field.

\paragraph{Multiscale SSM.} 
The standard Mamba SSM uses a data-dependent selection mechanism to generate its parameters, which is inadequate for capturing distinct autocorrelation patterns across scales. Using separate SSMs per scale increases parameter count linearly and fails to model cross-scale interactions, while sharing a single SSM causes information entanglement and degrades performance. To address this trade-off, we propose a novel multiscale SSM module that enables simultaneous multiscale pattern extraction.

We first concatenate temporal features from different scales along the feature dimension to form $\hat{h} \in \mathbb{R}^{T\times (Q \cdot d)}$, then project it to form the input-aware parameters as follows:
\begin{equation}
    [\Delta, B, C] = \text{tanh}(W_{\text{proj}}(\hat{h})).
\end{equation}
The input-aware parameters are learned across all scales, enabling a comprehensive understanding of the series. However, sharing them across all sequences ignores scale-specific differences. To address this, the multiscale SSM introduces a learnable bias term $\Omega \in \mathbb{R}^{(Q \cdot d)}$ added to $\Delta$, guiding the SSM to perform scale-specific pattern extraction. This modified $\Delta$ enables precise control over the SSM’s behavior by adjusting the discretization of matrices $A$ and $B$. The scale-specific and input-aware parameters are then updated as:
\begin{equation}
\hat{\Delta} = \text{Linear}({\Delta}) + \text{broadcast}(\Omega), ~~~~\tilde{A} = \exp(\hat{\Delta} \otimes A), ~~~~ \tilde{B} = \hat{\Delta} \otimes B.
\end{equation}
Based on the above parameters, the SSM function is introduced as:
\begin{equation}
    u^{(t)} = \tilde{A} u^{(t-1)} + \tilde{B} \hat{h}^{(t)}, ~~~~ h^{(t)}_{\text{final}} = C u^{(t)}.
\end{equation}
By integrating the scale amplification with the Multiscale SSM, Multiscale Mamba effectively captures multiscale temporal patterns.

\begin{algorithm}[t]
\fontsize{7.5pt}{8.5pt}\selectfont % 字号，行距

\caption{Multiscale Mamba}
\label{alg:multiscale_mamba}
\begin{algorithmic}
\REQUIRE Input sequence $H_{\text{in}} \in \mathbb{R}^{T \times d \times Q}$, scales $\mathcal{S} = [s_1,...,s_Q]\in \mathbb{R}^{Q}$
\ENSURE Output sequence $H_{\text{out}} \in \mathbb{R}^{T \times d \times Q}$

\STATE \textbf{1. Input Projection}
\STATE $\hat{H_{\text{in}}} \gets \text{Concat}(H_{\text{in}}), \hat{H_{\text{in}}} \in \mathbb{R}^{T \times (d \cdot Q)}$
\STATE $[h, z] \gets \text{Linear}(\hat{H_{\text{in}}})$, $h \in \mathbb{R}^{T \times (d_{\text{inner}} \cdot Q)}$, $z \in \mathbb{R}^{T \times (d_{\text{inner}} \cdot Q)}$

\STATE \textbf{2. Multiscale Feature Extraction}
\STATE $h_{\text{ms}} = \text{reshape}(h) \in \mathbb{R}^{T \times d_{\text{inner}} \times Q}$
\FOR{$q \gets 1$ \TO $Q$}
    \STATE $h^{(q)} \gets \text{Conv1D}(h_{\text{ms}}^{(q)}, s_q)$
\ENDFOR
\STATE $\hat{h} \gets \text{concat}(h^{(1)},...,h^{(Q)})$, $\hat{h} \in \mathbb{R}^{T \times (d_{\text{inner}} \cdot Q)}$

\STATE \textbf{3. Global Temporal Mixing}
\STATE $\hat{h} \gets \text{SiLU}(\text{CausalConv1D}(\hat{h}))$

\STATE \textbf{4. Selection Mechanism With Multiscale Bias}
\STATE $[\Delta, B, C] \gets \text{tanh}(W_{\text{proj}}(\hat{h}))$
\STATE $\hat{\Delta} \gets \text{Linear}({\Delta}) + \text{broadcast}(\Omega)$, $\hat{\Delta} \in \mathbb{R}^{T \times (d_{\text{inner}} \cdot Q)}$, $\Omega \in \mathbb{R}^{(d_{\text{inner}} \cdot Q)}$

\STATE \textbf{5. Discretization}
\STATE $\tilde{A} \gets \exp(\hat{\Delta} \otimes A)$
\STATE $\tilde{B} \gets \hat{\Delta} \otimes B$

\STATE \textbf{6. Parallel Selective Scan}
\STATE $h_{\text{final}} \gets \text{SSM}(\tilde{A}, \tilde{B}, C, \hat{h})$, $h^{\text{final}} \in \mathbb{R}^{T \times (d_{\text{inner}} \cdot Q)}$

\STATE \textbf{7. Gated Output}
\STATE $h_{\text{out}} \gets \text{SiLU}(z) \odot h_{\text{final}}$, $h^{\text{out}} \in \mathbb{R}^{T \times (d_{\text{inner}} \cdot Q)}$
\STATE $\hat{H_{\text{out}}} \gets \text{Linear}(h_{\text{out}}), \hat{H_{\text{out}}} \in \mathbb{R}^{T \times (d \cdot Q)}$
\STATE $H_{\text{out}} \gets \text{Split}(\hat{H_{\text{out}}})$

\RETURN $H_{\text{out}}$
\end{algorithmic}
\end{algorithm}

\subsection{Disentangled Mixture of Multiscale Mamba}
We have now introduced Multiscale Mamba for efficient multiscale information extraction and AGCCN for learning complex multiscale spatial-temporal correlations. \textbf{However, a single Multiscale Mamba struggles to capture diverse temporal dynamics as spatial-temporal dependencies become increasingly complex.} Therefore, we propose Disentangled Mixture of Multiscale Mamba (DMMM), which utilizes Disentangled Mixture of Experts (DMoE) mechanism with the Multiscale Mamba serving as experts.

\paragraph{Experts.} In the DMMM layer, we implement two types of experts: (1) a Multiscale Mamba serving as the shared expert $E_{0}$ that processes all input data, and (2) $K$ Multiscale Mamba with distinct parameters as specialized experts $\{E_k\}_{k=1}^K$ that process specified input data. Each expert $E_k(\cdot): \mathbb{R}^{T \times d \times Q } \rightarrow \mathbb{R}^{T \times d \times Q }, k \in [0, K]$ learns different temporal patterns. Since temporal sequences typically contain both shared and unique multiscale patterns, the shared expert $E_0$ extracts common patterns while the specialized experts $\{E_k\}_{k=1}^K$ focus on distinctive temporal patterns.

\begin{figure}[t]
% \begin{wrapfigure}{r}{0.5\textwidth}
    % \vspace{-10pt}
    \centering
    \includegraphics[width=\columnwidth]{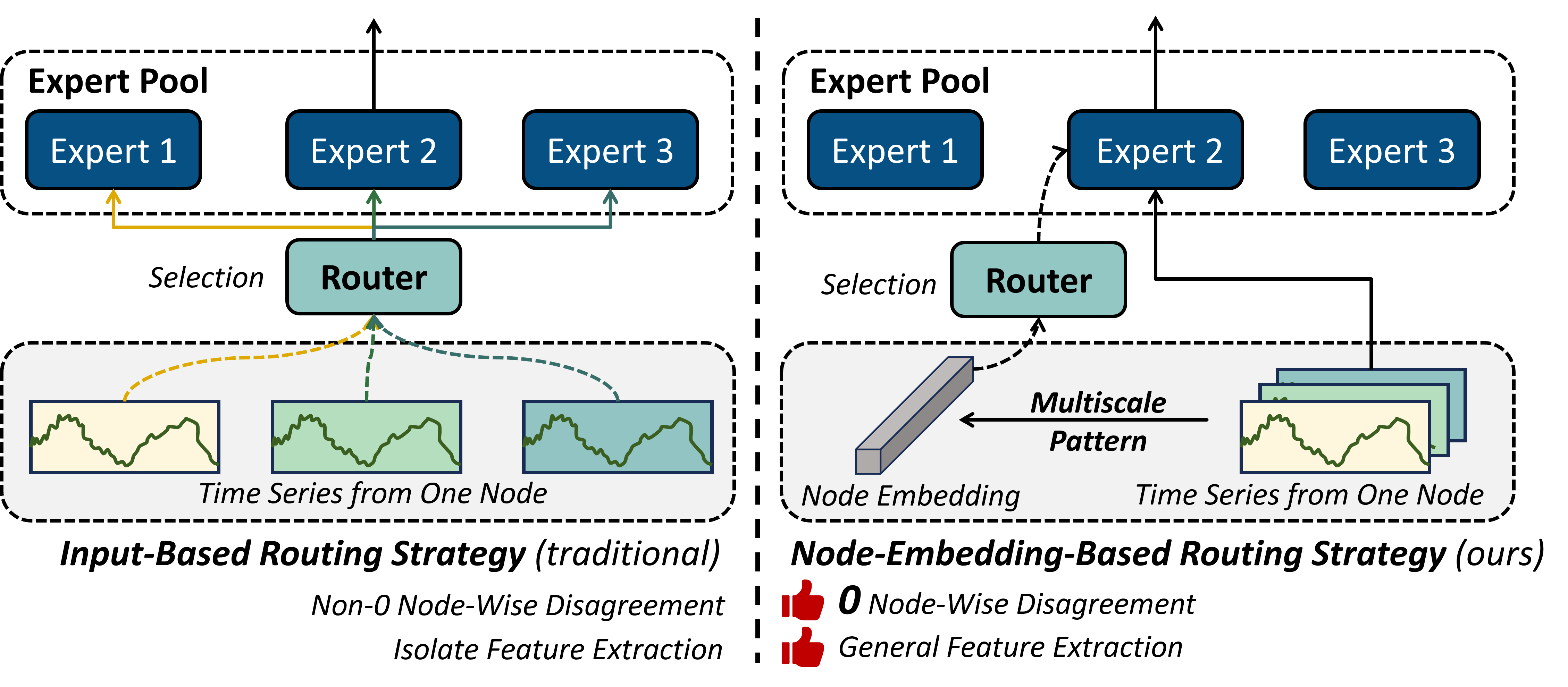}
    \caption{Comparison of routing strategies. (Left) Traditional Input-Based Strategy causes inconsistent expert assignment for time-series steps in the same node (non-zero node-wise disagreement). (Right) Our Node-Embedding-Based Strategy ensures zero node-wise disagreement via node-specified embedding, facilitating stable and specialized feature extraction.}
    \label{fig:routing}
    \vspace{-10pt}
% \end{wrapfigure}
\end{figure}

\paragraph{Routing Strategy.} We propose a novel \textbf{node-embedding-based routing strategy} to determine the appropriate specialized expert. Unlike traditional approaches that directly use input series for gating decisions, we employ the learnable node embeddings $E_A$ introduced in Section~\ref{sec:AGCCN}. These embeddings serve as node identifiers and provide more stable routing patterns. Formally, we define the gating network as $G(\cdot): \mathbb{R}^{d_e} \rightarrow \mathbb{R}^{K}$, implemented as a bias-free linear layer with added random noise to improve routing robustness \cite{chen2022towards}. For an input time series $x_{\text{in}} \in \mathbb{R}^{T \times d \times Q }$ of a node with corresponding learnable embedding $e_{\text{in}} \in \mathbb{R}^{d_e}$ capturing its intrinsic characteristics, the DMMM output $z \in \mathbb{R}^{T \times d \times Q }$ is computed as:
\begin{equation}
z = \textstyle\sum_{k=1}^{K}\mathbb{I}[i=g]E_k(x_{\text{in}}) + E_0(x_{\text{in}}),~~
g = {\mathrm{argmax}}_{k\in[K]} \{G_k(e_{in}) + r_{k}\},
\end{equation}
where $g$ denotes the top-1 selection of experts, $G_k$ denotes the gating network's output for expert $k$, $r$ denotes a random noise $r \sim \mathcal{U}(0, b_{up})$. We compare the traditional input-based routing strategy with ours in Figure~\ref{fig:routing}, and show that our routing strategy exhibits smoother behavior by Lemma~\ref{lem:pairwise_routing_disagreement}.

\begin{lemma}\label{lem:pairwise_routing_disagreement}
Let $\mathbf{H}=[\mathbf{h}_1, \mathbf{h}_2, ..., \mathbf{h}_M] \in \mathbb{R}^{M\times K}$ be the set of the gating network's output of $M$ samples, which belongs to one spatial node and $\mathbf{r}=[r_1, r_2, ..., r_K] \in \mathbb{R}^{K}$ be independent noise vectors drawn from $\mathcal{D}_r$ with $\rho$-bounded probability density function. Define the pairwise routing disagreement probability $p_{i,j}$ and the node-wise disagreement probability $\mathbf{P}$ as follows:
\begin{equation}
p_{i,j} = \mathbb{P}({\mathrm{argmax}}_{k\in[K]}\{h_{i,k} + r_{k}\} \neq {\mathrm{argmax}}_{k\in[K]} \{h_{j,k} + r_{k}\}),
\end{equation}
\begin{equation}
\begin{aligned}
\mathbf{P} = \mathbb{P}(\exists\, i \neq j \in [M], \text{ s.t. } 
&\; {\mathrm{argmax}}_{k\in[K]} \{h_{i,k} + r_{k}\} \\
&\neq {\mathrm{argmax}}_{k\in[K]} \{h_{j,k} + r_{k}\}).
\end{aligned}
\end{equation}
Then it satisfies:
\begin{equation}
0 \leq \mathbf{P} \leq \rho K^2 \textstyle\sum_{i < j} \|\mathbf{h}_i - \mathbf{h}_j\|_\infty,
\end{equation}
and for our node-embedding-based routing, $\mathbf{P} \equiv 0$.
\end{lemma}

Lemma~\ref{lem:pairwise_routing_disagreement} shows that our node-embedding-based routing achieves $\mathbf{P} \equiv 0$, satisfying the key design principle: sequence samples from the same spatial nodes share identical transition patterns, and ensuring smooth routing. In contrast, conventional input-based routing always produces non-zero $\mathbf{P}$, resulting in routing unsmoothness. The complete proof is provided in Appendix~\ref{appendix:proof:gating}.

\paragraph{Loss Design.} 
STM3 can predict the future data $\{\tilde{x}_{t+i}\}^{\tau}_{i=1}$. Initially, we can construct the MAE loss $\mathcal{L}_{MAE} = \mathbb{E}_{i=1}^\tau \|\tilde{x}_{t+i}-x_{t+i}\|^2_2$.
However, directly optimizing $\mathcal{L}_{MAE}$ to learn from specialized experts may lead to unstable training. If the gating network fails to provide stable routing, each expert may extract similar information.

To encourage experts to learn distinct content, contrastive learning offers an intuitive solution: treating outputs from the same expert as positive samples and those from different experts as negative samples. However, it assumes correct routing during training. In practice, we cannot guarantee that two samples routed to the same expert are truly similar when routing errors occur. Thus, the challenge lies in identifying reliable positive samples.

Fortunately, our analysis shows that coarser-scale outputs within a sample can effectively serve as positive samples for learning higher-level features, while finer-scale outputs should be treated as negatives to preserve scale distinguishability. This leads to an asymmetric similarity measure where $s(x, x') \neq s(x', x)$. Traditional contrastive learning relies on symmetric metrics (e.g., cosine similarity), which are unsuitable in this context. To address this, we propose a novel \textbf{causal contrastive learning} method.
First, we define an asymmetric causal measurement: for two outputs $(x^{(p)}_i, x^{(q)}_j)$ at scales $p$ and $q$, the similarity score is computed as:
\begin{equation}
	s(x^{(p)}_i,x^{(q)}_j) =\left\{
	\begin{aligned}
		&|p-q+1|^{-\gamma_1}\cos(x^{(p)}_i,x^{(q)}_j), ~~ \text{if $p > q$};  \\
		&|q-p+1|^{\gamma_2} \cos(x^{(p)}_i,x^{(q)}_j), ~~~\text{if $p \le q$},
	\end{aligned}
	\right.
\end{equation}
where $\gamma_1$ and $\gamma_2$ are non-negative hyperparameters that encourage feature diversity among experts. The causal similarity measure assigns higher similarity to positive pairs (coarser to finer scales), promoting alignment of samples with small-scale gaps. For negative pairs (finer to coarser), it assigns lower similarity to push apart samples with larger scale gaps.

For expert $k$ in our causal contrastive learning framework, consider the DMMM output $z^{p}_{i,k}$ from sample $i \in \mathcal{B}_k$ at scale $p$. As mentioned previously, while outputs $z^{p}_{j,k}$ from other samples $j \in \mathcal{B}_k$ at the same scale serve as natural positive candidates, the sample $j$ has the risk of being routed incorrectly. To mitigate this, we augment the positive set with higher-scale representations ($q > p$) of the same sample. The negative set comprises both lower-scale representations and outputs from other experts. Formally, we define the positive and negative sets for $z^{p}_{i,k}$ as $\mathcal{P}^{p}_{i,k}$ and $\mathcal{N}^{p}_{i,k}$ respectively, yielding the following contrastive loss:
\begin{equation}
\begin{aligned}
\mathcal{L}_{\text{C}}(k) = & - \textstyle\sum_{i\in \mathcal{B}_k} \sum^Q_{p=1} \frac{1}{|\mathcal{P}^{p}_{i,k}|}\sum_{z^*\in \mathcal{P}^{p}_{i,k}} \\
& \log \frac{\exp\big(s(z^{(p)}_{i,k},z^*) \cdot \theta\big) }{\sum_{z^{'}\in \mathcal{N}^{p}_{i,k}\cup \{z^*\}} \exp\big(s(z^{(p)}_{i,k},z^{'}) \cdot \theta\big)}.
\end{aligned}
\end{equation}
We use $\mathcal{L}_{\text{C}}(k)$ as an auxiliary loss to update the DMMM layer and construct the final loss $\mathcal{L} = \mathcal{L}_{\text{MAE}} + \lambda \textstyle\sum_{l=1}^L \mathbb{E}_{k=1}^K (\mathcal{L}_{\text{C}}(k))$, where $L$ denotes the number of backbone layers, and $\lambda$ is a hyperparameter.

\paragraph{Pattern Disentanglement for DMMM.}
Each expert in DMMM should focus on its respective temporal patterns, which essentially aim at achieving pattern disentanglement. Next, we explain in detail how $\mathcal{L}_{\text{C}}(k)$ achieves cross-expert pattern disentanglement, theoretically inspired by IP-IRM \cite{wang2021self}. Assuming feature space $\mathcal{X}$ is homogeneous under group $G$, any feature $x' \in \mathcal{X}$ can be transformed from $x \in \mathcal{X}$ via group action $g \cdot x$. We define the key concepts:

\begin{definition}[Orbit and Partition]
Given a subgroup $D \subset G$, it partitions the feature space $\mathcal{X}$ into $k$ disjoint subsets as $\mathcal{X} = \textstyle\cup_{i=1}^k D(c_i \cdot x)$, where $\{c_i D\}^{k}_{i=1}$ are the cosets. These cosets form a factor group $G/D = \{c_i\}_{i=1}^k$. Here, each $c_i \cdot x$ serves as a representative feature for the $i$-th class, obtained through the group action on any feature $x \in \mathcal{X}$.
\end{definition}

% DMMM's top-1 routing strategy partitions input samples into $k$ distinct classes, where $G/D$ denotes feature transformations across experts and $D$ denotes feature transformations within one expert. We consider a well-disentangled representation that can be separated into a \emph{class-agnostic} component, which is invariant to $G/D$ (cross-orbit) and equivariant to $D$ (in-orbit), and a \emph{class-specific} component, which is equivariant to $G/D$ and invariant to $D$. We now prove that such disentanglement can be achieved through contrastive learning:
DMMM’s top-1 routing strategy partitions input samples into $k$ distinct classes, where $G/D$ denotes transformations across experts, and $D$ denotes transformations within an expert. We define a well-disentangled representation that can be separated into a \emph{class-agnostic} component, which is invariant to $G/D$ (cross-orbit) and equivariant to $D$ (in-orbit), and a \emph{class-specific} component, which is equivariant to $G/D$ and invariant to $D$. We now prove that such disentanglement can be achieved via contrastive learning:

% In our framework, each training feature $x$ is the multiscale views generated from the multiscale transformation $D$. We consider a well-disentangled representation as one that can be separated into a \emph{class-agnostic} component, which is invariant to $G/D$ (cross-orbit) and equivariant to $D$ (in-orbit), and a \emph{class-specific} component, which is equivariant to $G/D$ and invariant to $D$. This enables accurate prediction by retaining only the class-specific component. We now prove that such disentanglement can be achieved through contrastive learning:

\begin{lemma}\label{lemma1}
% The training loss $-\log\frac{\exp(s(x_i, x_j))}{\sum_{x \in \mathcal{X}} \exp(s(x_i,x))}$ disentangles $\mathcal{X}$ with respect to $(G/D) \times D$, where $x_i$ and $x_j$ are from the same orbit.
The contrastive loss $-\log\frac{\exp(s(x_i, x_j))}{\sum_{x \in \mathcal{X}} \exp(s(x_i,x))}$ enforces the disentanglement of feature space $\mathcal{X}$ for $(G/D) \times D$, where $(x_i, x_j)$ are from the same orbit under group action $D$.
\end{lemma}

% The complete proof is provided in the Appendix \ref{appendix:proof:disentangle}. This lemma establishes a theoretical connection between contrastive learning and disentangled representation learning. Specifically, when we construct positive feature pairs using the features related via the group action $D$, while treating all others as negative features, we implicitly encourage disentanglement with respect to $D$. Although Lemma~\ref{lemma1} does not guarantee a complete decomposition for every element $d \in D$, the downstream model can still leverage prediction-specific features associated with $G/D$.
The complete proof is provided in the Appendix \ref{appendix:proof:disentangle}. This lemma establishes a theoretical connection between contrastive learning and disentangled representation learning. In the single-scale case, our contrastive loss $\mathcal{L}_{\text{C}}(k)$ simplifies to the conventional contrastive loss while still preserving cross-expert pattern disentanglement. The multiscale extension preserves this disentanglement while additionally emphasizing cross-scale differences, thereby enhancing DMMM's multiscale processing capability.

\iffalse{In our framework, $D$ represents the multiscale view generation process, while the DMMM's top-1 routing strategy defines $\mathcal{X}$. The contrastive loss effectively decouples the pattern extraction across different experts, ensuring that each expert's output remains class-specific—capturing the shared characteristics of its input distribution.}\fi

\defcitealias{Hu_Liu_Zhu_Cheng_Dai_2025}{(AAAI 25)}
\defcitealias{Wang_Liu_Duan_Wang_2025}{(AAAI 25)}
\defcitealias{wang2025timemixer++}{(ICLR 25)}

\defcitealias{Zhou_Zhang_Peng_Zhang_Li_Xiong_Zhang_2021}{(AAAI 21)}
\defcitealias{zhang2023promptst}{(CIKM 23)}
\defcitealias{nie2023a}{(ICLR 23)}
\defcitealias{lablack2023spatio}{(ESWA 23)}
\defcitealias{liu2024itransformer}{(ICLR 24)}
\defcitealias{yan2025dgraformer}{(IJCAI 25)}

\defcitealias{xu2025sst}{(CIKM 25)}

\defcitealias{li2024stg}{(arXiv 24)}
\defcitealias{alkilane2024mixmamba}{(IF 24)}
\defcitealias{He_Ji_Lei_2025}{(AAAI 25)}

\begin{table*}[htbp]
  \centering
  \caption{Prediction results with input-output steps $T-\tau = 12-\{12, 24, 48, 96\}$ for Milan and $T-\tau = 96-\{96, 192, 336, 720\}$ for others. Results are averaged from all prediction lengths. The best performance is highlighted in \textbf{\textcolor{red}{red}}, and the second-best is \underline{\textcolor{blue}{underlined}}. Full results are listed in Appendix~\ref{appendix:full_compare}.}
  \label{tab:results1}
  \resizebox{\textwidth}{!}{%
    \setlength{\tabcolsep}{3.4pt} % 列间距，默认是6pt
    \fontsize{10pt}{11pt}\selectfont % 字号，行距
    
    \begin{tabular}{lccccccccccc}
    \toprule
    \multirow{2}{*}{\textbf{Model}} & \multicolumn{3}{c}{\textbf{METR-LA}} & \multicolumn{3}{c}{\textbf{PEMSD4}} & \multicolumn{3}{c}{\textbf{PEMSD8}} & \multicolumn{2}{c}{\textbf{ETTh1}} \\
    \cmidrule(lr){2-4} \cmidrule(lr){5-7} \cmidrule(lr){8-10} \cmidrule(lr){11-12}
     & \textbf{MAE} & \textbf{RMSE} & \textbf{MAPE} & \textbf{MAE} & \textbf{RMSE} & \textbf{MAPE} & \textbf{MAE} & \textbf{RMSE} & \textbf{MAPE} & \textbf{MAE} & \textbf{MSE} \\
    \midrule
    Informer \citetalias{Zhou_Zhang_Peng_Zhang_Li_Xiong_Zhang_2021} & 14.5265 & 18.9395 & 0.2353 & 37.6445 & 52.4407 & 0.3363 & 33.5953 & 46.7593 & 0.2383 & 0.6895 & 0.8118 \\
    PromptST \citetalias{zhang2023promptst} & 12.6286 & 19.8336 & 0.2543 & 85.8462 & 108.8736 & 1.1915 & 75.6295 & 94.7616 & 0.8077 & 0.6523 & 0.8242 \\
    PatchTST \citetalias{nie2023a} & 14.1736 & 19.4963 & 0.2837 & 45.2831 & 65.9194 & 0.3808 & 45.2831 & 65.9194 & 0.3807 & 0.4355 & \underline{\textcolor{blue}{0.4419}} \\
    STGM \citetalias{lablack2023spatio} & 12.4934 & \underline{\textcolor{blue}{17.5122}} & 0.2316 & 45.0099 & 63.7933 & 0.3712 & 36.2963 & 50.5469 & 0.2880 & 0.4470 & 0.4590 \\
    iTransformer \citetalias{liu2024itransformer} & 13.8924 & 18.7139 & 0.2582 & 42.6569 & 61.2748 & 0.4395 & 37.2686 & 54.5294 & 0.3256 & 0.4409 & 0.4452 \\
    TimeMixer++ \citetalias{wang2025timemixer++} & 14.1992 & 20.2049 & 0.2562 & 36.9178 & 53.3192 & 0.3694 & 32.8147 & 47.2128 & 0.2697 & 0.7952 & 0.9755 \\
    AMD \citetalias{Hu_Liu_Zhu_Cheng_Dai_2025} & 13.8865 & 20.5249 & 0.2831 & 73.8647 & 101.9244 & 0.8743 & 63.9017 & 89.5960 & 0.5518 & 0.4616 & 0.4799 \\
    FilterTS \citetalias{Wang_Liu_Duan_Wang_2025} & 13.9973 & 19.2234 & 0.2648 & 54.9357 & 75.9463 & 0.5877 & 51.2711 & 70.7446 & 0.4107 & 0.4435 & 0.4525 \\
    DGraFormer \citetalias{yan2025dgraformer} & \underline{\textcolor{blue}{12.4590}} & 20.3517 & 0.2546 & 51.2170 & 73.8316 & 0.5639 & 42.1643 & 60.7884 & 0.3656 & \underline{\textcolor{blue}{0.4308}} & 0.4428 \\
    % \multicolumn{12}{c}{\textit{---------------------------- Mamba-based ----------------------------}} \\
    \cdashline{1-12}[1pt/1pt]
    STGMamba \citetalias{li2024stg} & 13.4377 & 18.4689 & \underline{\textcolor{blue}{0.2148}} & \underline{\textcolor{blue}{29.2433}} & \underline{\textcolor{blue}{44.7829}} & \underline{\textcolor{blue}{0.2309}} & 29.3542 & \underline{\textcolor{blue}{42.5652}} & \underline{\textcolor{blue}{0.2151}} & 0.6510 & 0.7302 \\
    MixMamba \citetalias{alkilane2024mixmamba} & 12.9707 & 18.8534 & 0.2492 & 62.7751 & 85.9823 & 0.7385 & 48.3085 & 68.1493 & 0.4113 & \textbf{\textcolor{red}{0.4257}} & \textbf{\textcolor{red}{0.4342}} \\
    SST \citetalias{xu2025sst} & 14.0114 & 19.8707 & 0.2842 & 60.1591 & 84.2085 & 0.6422 & 52.1986 & 73.0623 & 0.4296 & 0.4471 & 0.4581 \\
    DST-Mamba \citetalias{He_Ji_Lei_2025} & 13.8662 & 18.6325 & 0.2507 & 33.5377 & 49.7239 & 0.3294 & \underline{\textcolor{blue}{27.9594}} & 42.9988 & 0.2295 & 0.4478 & 0.4551 \\
    \textbf{\textit{STM3 (Ours)}} & \textbf{\textcolor{red}{10.9759}} & \textbf{\textcolor{red}{17.2195}} & \textbf{\textcolor{red}{0.1747}} & \textbf{\textcolor{red}{28.8715}} & \textbf{\textcolor{red}{43.4647}} & \textbf{\textcolor{red}{0.2214}} & \textbf{\textcolor{red}{25.9557}} & \textbf{\textcolor{red}{38.9235}} & \textbf{\textcolor{red}{0.1808}} & 0.4363 & 0.4467 \\
    \bottomrule
    \end{tabular}%
  }
\end{table*}
\begin{table*}[htbp]
  \centering
  % \caption{Performance comparison...}
  \label{tab:results2}
  \vspace{-10pt}
  
  \resizebox{\textwidth}{!}{%
    \setlength{\tabcolsep}{2.1pt} % 列间距，默认是6pt
    \fontsize{10pt}{12pt}\selectfont % 字号，行距
    \begin{tabular}{lcccccccccccc}
    \toprule
    \multirow{2}{*}{\textbf{Model}} & \multicolumn{2}{c}{\textbf{KnowAir}} & \multicolumn{2}{c}{\textbf{NREL}} & \multicolumn{2}{c}{\textbf{Milan-SMS}} & \multicolumn{2}{c}{\textbf{Milan-Call}} & \multicolumn{2}{c}{\textbf{Milan-Internet}} & \multicolumn{2}{c}{\textbf{Electricity}} \\
    \cmidrule(lr){2-3} \cmidrule(lr){4-5} \cmidrule(lr){6-7} \cmidrule(lr){8-9} \cmidrule(lr){10-11} \cmidrule(lr){12-13}
     & \textbf{MAE} & \textbf{RMSE} & \textbf{MAE} & \textbf{RMSE} & \textbf{MAE} & \textbf{RMSE} & \textbf{MAE} & \textbf{RMSE} & \textbf{MAE} & \textbf{RMSE} & \textbf{MAE} & \textbf{MSE} \\
    \midrule
    Informer \citetalias{Zhou_Zhang_Peng_Zhang_Li_Xiong_Zhang_2021} & 20.3528 & 26.7873 & 6.4849 & 8.1280 & 20.0953 & 45.7634 & 21.1082 & 46.3704 & 118.3482 & 283.7426 & 0.3840 & 0.2955 \\
    PromptST \citetalias{zhang2023promptst} & 17.9585 & 24.7702 & 9.7293 & 12.4912 & 21.2469 & 45.9126 & 20.4830 & 45.2762 & 95.9964 & 234.2947 & 0.6760 & 0.7979 \\
    PatchTST \citetalias{nie2023a} & 17.7040 & 24.7386 & 6.0000 & 8.1331 & 17.8521 & 40.5733 & 12.6166 & 33.4862 & 60.2950 & 156.8898 & 0.2752 & 0.1900 \\
    STGM \citetalias{lablack2023spatio} & \underline{\textcolor{blue}{17.2308}} & \underline{\textcolor{blue}{24.0108}} & 5.9728 & 7.8552 & 17.1129 & 36.3073 & 13.8053 & 32.6318 & 74.4659 & 173.5840 & 0.2900 & 0.2099 \\
    iTransformer \citetalias{liu2024itransformer} & 17.7803 & 24.4335 & 5.9325 & \underline{\textcolor{blue}{7.7270}} & \underline{\textcolor{blue}{15.8021}} & \underline{\textcolor{blue}{35.7936}} & \underline{\textcolor{blue}{12.3152}} & \underline{\textcolor{blue}{30.8416}} & 59.3740 & 154.5030 & \underline{\textcolor{blue}{0.2693}} & 0.1772 \\
    TimeMixer++ \citetalias{wang2025timemixer++} & 19.4032 & 26.2646 & 6.3777 & 8.5095 & 29.9546 & 54.7220 & 29.4299 & 52.6730 & 206.7419 & 365.5573 & 0.4603 & 0.4048 \\
    AMD \citetalias{Hu_Liu_Zhu_Cheng_Dai_2025} & 19.5950 & 30.0215 & 8.0247 & 10.8220 & 17.8179 & 39.5774 & 14.3319 & 35.9171 & 68.1947 & 165.1333 & 0.3167 & 0.2297 \\
    FilterTS \citetalias{Wang_Liu_Duan_Wang_2025} & 17.7152 & 24.3081 & 6.3209 & 8.3300 & 18.1332 & 40.9032 & 14.1871 & 35.6803 & 60.4592 & 156.0697 & 0.2867 & 0.1910 \\
    DGraFormer \citetalias{yan2025dgraformer} & 17.4259 & 24.9249 & \underline{\textcolor{blue}{5.9036}} & 8.1954 & 17.5319 & 41.1053 & 12.6293 & 34.0892 & \underline{\textcolor{blue}{57.7259}} & 151.8126 & 0.2802 & 0.1984 \\
    % \multicolumn{13}{c}{\textit{---------------------------- Mamba-based ----------------------------}} \\
    \cdashline{1-13}[1pt/1pt]
    STGMamba \citetalias{li2024stg} & 20.0842 & 26.3844 & 6.1707 & 7.9411 & 19.4521 & 40.6573 & 17.4995 & 38.0818 & 137.5390 & 275.5953 & 0.3960 & 0.3212 \\
    MixMamba \citetalias{alkilane2024mixmamba} & 17.4612 & 24.2597 & 5.9389 & 7.8114 & 16.6733 & 37.7126 & 12.5650 & 32.0000 & 59.0008 & 154.9912 & 0.3045 & 0.2180 \\
    SST \citetalias{xu2025sst} & 17.9216 & 25.0719 & 6.3056 & 8.5535 & 18.7187 & 42.2342 & 14.5447 & 36.4183 & 62.6000 & 160.9235 & 0.2818 & 0.1996 \\
    DST-Mamba \citetalias{He_Ji_Lei_2025} & 17.8381 & 24.3403 & 5.9604 & 7.8124 & 17.9537 & 39.2982 & 13.8850 & 33.6094 & 58.3534 & \underline{\textcolor{blue}{150.3015}} & 0.2737 & \underline{\textcolor{blue}{0.1758}} \\
    \textbf{\textit{STM3 (Ours)}} & \textbf{\textcolor{red}{17.1012}} & \textbf{\textcolor{red}{23.9213}} & \textbf{\textcolor{red}{5.3064}} & \textbf{\textcolor{red}{7.1725}} & \textbf{\textcolor{red}{14.8341}} & \textbf{\textcolor{red}{35.0077}} & \textbf{\textcolor{red}{11.4069}} & \textbf{\textcolor{red}{29.8023}} & \textbf{\textcolor{red}{56.8593}} & \textbf{\textcolor{red}{144.6816}} & \textbf{\textcolor{red}{0.2591}} & \textbf{\textcolor{red}{0.1683}} \\
    \bottomrule
    \end{tabular}%
  }
\end{table*}

\section{Experiment}\label{sec:experiment}
To evaluate the effectiveness of STM3, we answer the following five research questions:

\begin{itemize}[itemsep=0em, topsep=0em, leftmargin=1em]
    \item \textbf{RQ1}: How do our STM3 models perform on spatio-temporal time-series prediction tasks across diverse real-world domains, compared with existing baseline methods?
    \item \textbf{RQ2}: What are the individual contributions of each proposed module to the overall performance of the model?
    \item \textbf{RQ3}: How do different hyperparameter choices influence the predictive performance and stability of STM3?
    \item \textbf{RQ4}: How does the proposed Disentangled Multiscale Mamba Module (DMMM) effectively extract disentangled temporal patterns for each expert at each scale?
    \item \textbf{RQ5}: How does our proposed node-embedding-based routing strategy improve expert assignment smoothness and downstream prediction accuracy?
\end{itemize}

\subsection{Experimental Setting}\label{sec:experimental_setting}

\paragraph{Datasets.} 
To comprehensively evaluate the effectiveness of STM3, we select several real-world datasets spanning diverse application domains. These include three widely used transportation datasets, METR-LA~\cite{li2018diffusion}, PEMSD4, and PEMSD8~\cite{chen2001freeway}, an air quality dataset KnowAir~\cite{wang2020pm2}, a wireless traffic dataset Milan~\cite{barlacchi2015multi} (with three types of record), a solar power dataset NREL~\cite{wu2021inductive}, and two electricity datasets, ETTh1 and Electricity~\cite{liu2024itransformer}. These datasets have been extensively adopted in prior spatio-temporal time-series prediction research.
The dataset was split into training, validation, and testing parts with a 6:2:2 ratio for METR-LA, PEMSD4, PEMSD8, KnowAir, NREL, and Milan datasets. The split for the ETTh1 (8545, 2881, 2881) and Electricity (18317, 2633, 5261) datasets is similar to the common setting used in previous time-series prediction works~\cite{liu2024itransformer, wang2025timemixer++}.

\begin{table*}[t]
\centering
\caption{Ablation study results (MAE) average over all in-out prediction settings. \textbf{Imp.} denotes the average performance improvement of STM3 compared to the variant. Full results are listed in Appendix~\ref{appendix:full_ablation}.}
\label{tab:ablation_summary}
\resizebox{\textwidth}{!}{
\setlength{\tabcolsep}{1.7pt} % 列间距，默认是6pt
\fontsize{10pt}{11pt}\selectfont % 字号，行距
\begin{tabular}{lccccccccccc}
\toprule
\textbf{Model} & \textbf{METR\_LA} & \textbf{PEMSD4} & \textbf{PEMSD8} & \textbf{KnowAir} & \textbf{NREL} & \textbf{ETTh1} & \textbf{Electricity} & \textbf{Milan-SMS} & \textbf{Milan-Call} & \textbf{Milan-Internet} & \textbf{Imp.} \\
\midrule
\rowcolor{rowgray}
\textbf{STM3} & \textbf{10.9759} & \textbf{28.8715} & \textbf{25.9557} & \textbf{17.1012} & \textbf{5.3064} & \textbf{0.4363} & \textbf{0.2591} & \textbf{14.8341} & \textbf{11.4069} & \textbf{56.8593} & -- \\
\hspace{1em} w/o DMoE & 11.1693 & 30.6038 & 28.9057 & 17.2749 & 5.3579 & 0.4957 & 0.2622 & 15.1443 & 12.6439 & 71.1947 & +6.47\% \\
\hspace{1em} w/o CL & 11.0391 & 29.9519 & 27.3244 & 17.2096 & 5.5039 & 0.4978 & 0.2617 & 15.0654 & 12.1965 & 66.5934 & +4.94\% \\
\hspace{1em} w/o Multiscale & 11.3852 & 34.0472 & 31.1776 & 17.1564 & 5.4570 & 0.4707 & 0.2708 & 15.6223 & 13.2818 & 78.6876 & +9.72\% \\
\hspace{1em} w/o CA & 11.2085 & 29.6878 & 27.0396 & 17.2407 & 5.5758 & 0.4866 & 0.2669 & 15.5310 & 12.4583 & 67.8722 & +5.69\% \\
\hspace{1em} vanilla Mamba & 11.1024 & 30.0680 & 28.7001 & 17.2296 & 5.4324 & 0.4643 & 0.2656 & 15.2603 & 12.6588 & 71.7852 & +5.97\% \\
\hspace{1em} vanilla routing & 11.1869 & 29.2622 & 27.3737 & 17.3007 & 5.4487 & 0.4527 & 0.2634 & 15.0323 & 12.4584 & 68.7830 & +4.45\% \\
\bottomrule
\end{tabular}}
\vspace{-10pt}
\end{table*}

\paragraph{Baselines and Metrics.} To assess the effectiveness of STM3, 13 advanced baselines are selected, including Mamba-based models: DST-Mamba\cite{He_Ji_Lei_2025},  SST\cite{xu2025sst}, MixMamba~\cite{alkilane2024mixmamba}, STGMamba~\cite{li2024stg}, and other types' models: DGraFormer~\cite{yan2025dgraformer}, FilterTS~\cite{Wang_Liu_Duan_Wang_2025}, AMD~\cite{Hu_Liu_Zhu_Cheng_Dai_2025}, TimeMixer++~\cite{wang2025timemixer++}, iTransformer~\cite{liu2024itransformer}, STGM~\cite{lablack2023spatio}, PatchTST~\cite{nie2023a}, PromptST\cite{zhang2023promptst}, Informer\cite{Zhou_Zhang_Peng_Zhang_Li_Xiong_Zhang_2021}. We employ common metrics: Mean Absolute Error (MAE), Root Mean Squared Error (RMSE), Mean Absolute Percentage Error (MAPE), and Mean Squared Error (MSE).

\paragraph{Implementation Details.} The experiments are conducted on 4 Nvidia RTX3090 GPUs and implemented with PyTorch. We used the AdamW optimizer with an initial learning rate of 3e-3. The batch size is 64, and the max training epochs are 100. The learning rate is halved after 25 epochs of training, and the early stop patience is 15 epochs. To evaluate STM3 under long-term temporal settings, we set the input-output step $T-\tau = 96-\{96, 192, 336, 720\}$ for METR-LA, PEMSD4, PEMSD8, KnowAir, NREL, ETTh1, and Electricity datasets. Due to the small size of the Milan dataset, we use $T -\tau = 12-\{12, 24, 48, 96\}$ for it, which is longer than the previous works' common setting \cite{yao2021mvstgn, mehrabian2023dynamic, liu2024spatial, mehrabian2025gamba}.

More detailed experimental settings are provided in Appendix~\ref{appendix:implement_detail}.

\subsection{Main Results (RQ1)}\label{sec:main_Results}

Table~\ref{tab:results1} demonstrates that \textbf{STM3 achieves state-of-the-art performance in long-term prediction across various datasets}, against Mamba-based and other types of approaches. For instance, on PEMSD8, it surpasses the second-best model by \textbf{7.1\%} in MAE, \textbf{8.5\%} in RMSE, and \textbf{15.9\%} in MAPE. On NREL, it surpasses the second-best model by \textbf{10.1\%} in MAE and \textbf{7.1\%} in RMSE.
This superior performance highlights the effectiveness of STM3’s architectural design.
Specifically, STM3 surpasses DST-Mamba~\cite{He_Ji_Lei_2025}, demonstrating that disentangling complex multiscale information is more effective than simple aggregation.

Regarding computational efficiency (detailed in Appendix~\ref{appendix:computation_cost}), although STM3 processes extensive multi-scale information through a multi-expert design and contrastive learning, which leads to an increase in training cost, it is a justified trade-off for its superior accuracy. STM3 remains significantly more efficient than complex baselines like AMD and MixMamba. By leveraging the cross-scale shared Mamba design to prevent parameter explosion and the Top-1 MoE mechanism for rapid inference, STM3 strikes a balance between superior performance and operational feasibility.

\begin{figure}[t]
    \centering
    
    \begin{minipage}[c]{0.49\columnwidth}
        \centering
        \includegraphics[height=2.5cm]{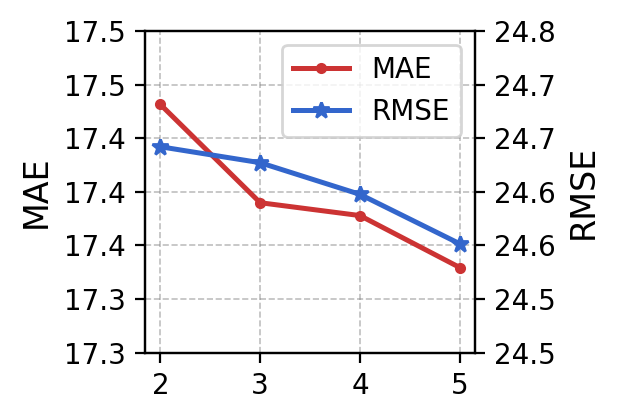}
        \subcaption{$K$ on KnowAir.}
        % \label{fig:}
    \end{minipage}
    \hfill
    \begin{minipage}[c]{0.49\columnwidth}
        \centering
        \includegraphics[height=2.5cm]{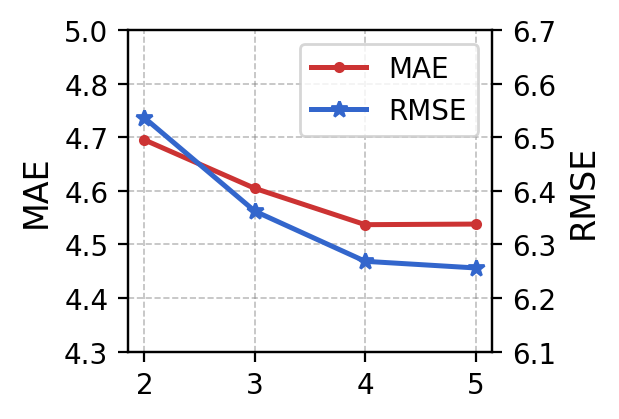}
        \subcaption{$K$ on NREL.}
        % \label{fig:}
    \end{minipage}
    \hfill
    \begin{minipage}[c]{0.49\columnwidth}
        \centering
        \includegraphics[height=2.5cm]{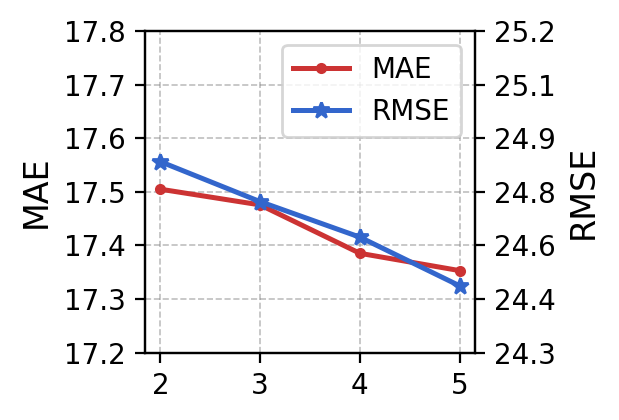}
        \subcaption{$Q$ on KnowAir.}
        % \label{fig:}
    \end{minipage}
    \hfill
    \begin{minipage}[c]{0.49\columnwidth}
        \centering
        \includegraphics[height=2.5cm]{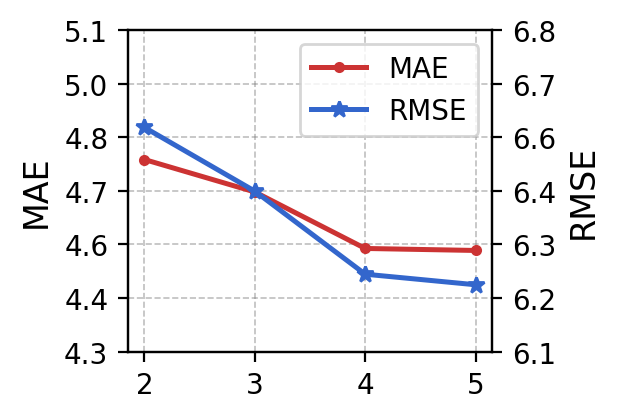}
        \subcaption{$Q$ on NREL.}
        % \label{fig:}
    \end{minipage}
    \hfill
    \begin{minipage}[c]{0.49\columnwidth}
        \centering
        \includegraphics[height=2.5cm]{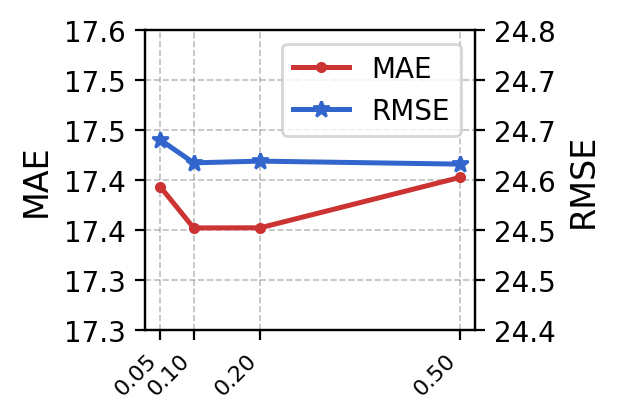}
        \subcaption{$\lambda$ on KnowAir.}
        % \label{fig:}
    \end{minipage}
    \hfill
    \begin{minipage}[c]{0.49\columnwidth}
        \centering
        \includegraphics[height=2.5cm]{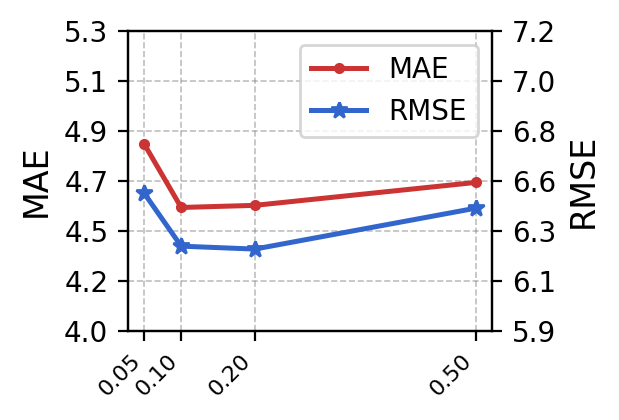}
        \subcaption{$\lambda$ on NREL.}
        % \label{fig:}
    \end{minipage}
    
    \caption{Hyperparameter study.}
    \label{fig:main_hyper}
    \vspace{-20pt}
\end{figure}

\subsection{Ablation Study (RQ2)}\label{sec:ablation}

To validate STM3's design, we conduct ablation studies on six variants: 1) \textbf{w/o DMoE}: only a single Multiscale Mamba expert is employed; 2) \textbf{w/o CL}: no disentangled contrastive learning; 3) \textbf{w/o Multiscale}: removes multiscale extraction; 4) \textbf{w/o CA}: no causal attention in AGCCN; 5) \textbf{vanilla Mamba}: replaces Multiscale Mamba with standard vanilla Mamba; 6) \textbf{vanilla routing}: input-based routing instead of node embeddings. We record the average MAE results of all in-out prediction settings in Table~\ref{tab:ablation_summary}. 
Multiscale mechanism yields the most substantial impact, providing an average performance gain of \textbf{9.72\%} (\textit{w/o Multiscale}). This aligns with our design principle that capturing temporal patterns at multiple scales is critical for accurate forecasting.
The DMoE design is the second most influential factor. Comparing STM3 with the single-model variant (\textit{w/o DMoE}), we observe a performance improvement of \textbf{6.47\%}, confirming that disentangling complex patterns effectively enhances model capacity.
Replacing our specialized Multiscale Mamba with a standard \textit{vanilla Mamba} leads to a \textbf{5.97\%} drop in performance, validating the effectiveness of this architecture design.
While individual contributions vary across datasets, the positive impact of each component confirms that they are indispensable for achieving optimal spatio-temporal prediction.
We provided the detailed ablation study results in Appendix~\ref{appendix:full_ablation}.

\subsection{Hyperparameter Study (RQ3)}\label{sec:hyperparameter}
% To investigate the impact of key hyperparameters on the performance of our proposed STM3, we conduct an analysis on KnowAir and NREL datasets, focusing on the three important hyperparameters: the number of specialized experts $K$, the number of scales $Q$, and the weight coefficient of disentangle contrastive learning loss $\lambda$. The results are illustrated in Fig.~\ref{fig:main_hyper}.
% (1) As the number of experts $K$ increases, the MAE and RMSE generally decrease across both datasets. This suggests that a larger number of experts provides the model with sufficient capacity to capture diverse and complex spatio-temporal patterns. While considering the cost-performance tradeoff, we suggest selecting $K=4$.
% (2) The prediction error generally decreases as the number of scales $Q$ increases. This validates the necessity of extracting multiscale patterns. While the performance gain may become marginal when $Q$ increases, indicating a potential saturation in feature representation. We suggest setting $Q$ from 3 to 4.
% (3) The error metrics exhibit a U-shaped pattern as $\lambda$ increases. The model achieves optimal performance when $\lambda$ is set around $0.1$ to $0.2$. A value of $\lambda$ that is too small may fail to effectively enforce decoupled representations, while an excessively large value may dominate the total loss, thereby distracting the optimization of the primary forecasting task.

We analyze the sensitivity of STM3 to three key hyperparameters ($K$, $Q$, $\lambda$) on the KnowAir and NREL datasets, as illustrated in Fig.~\ref{fig:main_hyper}. 
\textbf{(1) Number of Experts ($K$):} Increasing $K$ generally reduces errors, confirming that sufficient expert capacity is crucial for capturing diverse patterns. We select $K=4$ as the optimal cost-performance tradeoff. 
\textbf{(2) Number of Scales ($Q$):} Performance improves with higher $Q$, validating the benefit of multiscale modeling. However, since gains saturate at higher values, we recommend setting $Q \in [3, 4]$. 
\textbf{(3) Contrastive Weight ($\lambda$):} A U-shaped trend is observed, with optimal performance around $\lambda \in [0.1, 0.2]$. Extremes degrade results: low $\lambda$ fails to enforce disentanglement, while high $\lambda$ overshadows the primary forecasting loss.

\subsection{In-Depth Analysis (RQ4 \& RQ5)}\label{sec:in_depth_alalysis}

\begin{figure}[t]
% \begin{wrapfigure}{r}{0.63\textwidth}
    % \vspace{-40pt}
    \centering
    \begin{minipage}[c]{0.35\columnwidth}
        \centering
        \includegraphics[height=3.3cm]{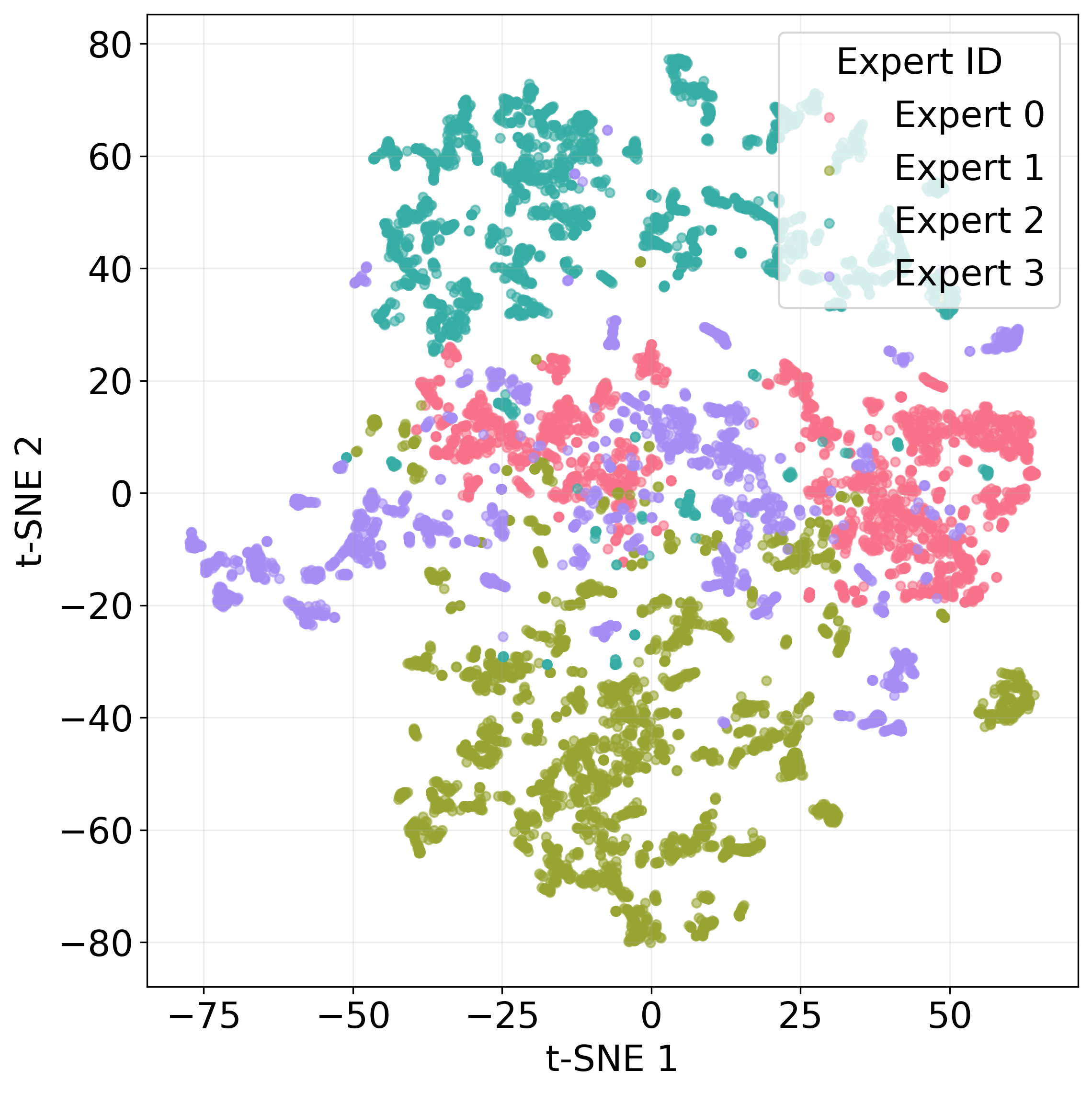}
        \subcaption{Expert-wise features.}
        \label{fig:moe_visulization}
    \end{minipage}
    \hfill
    \begin{minipage}[c]{0.62\columnwidth}
        \centering
        \includegraphics[height=3.3cm]{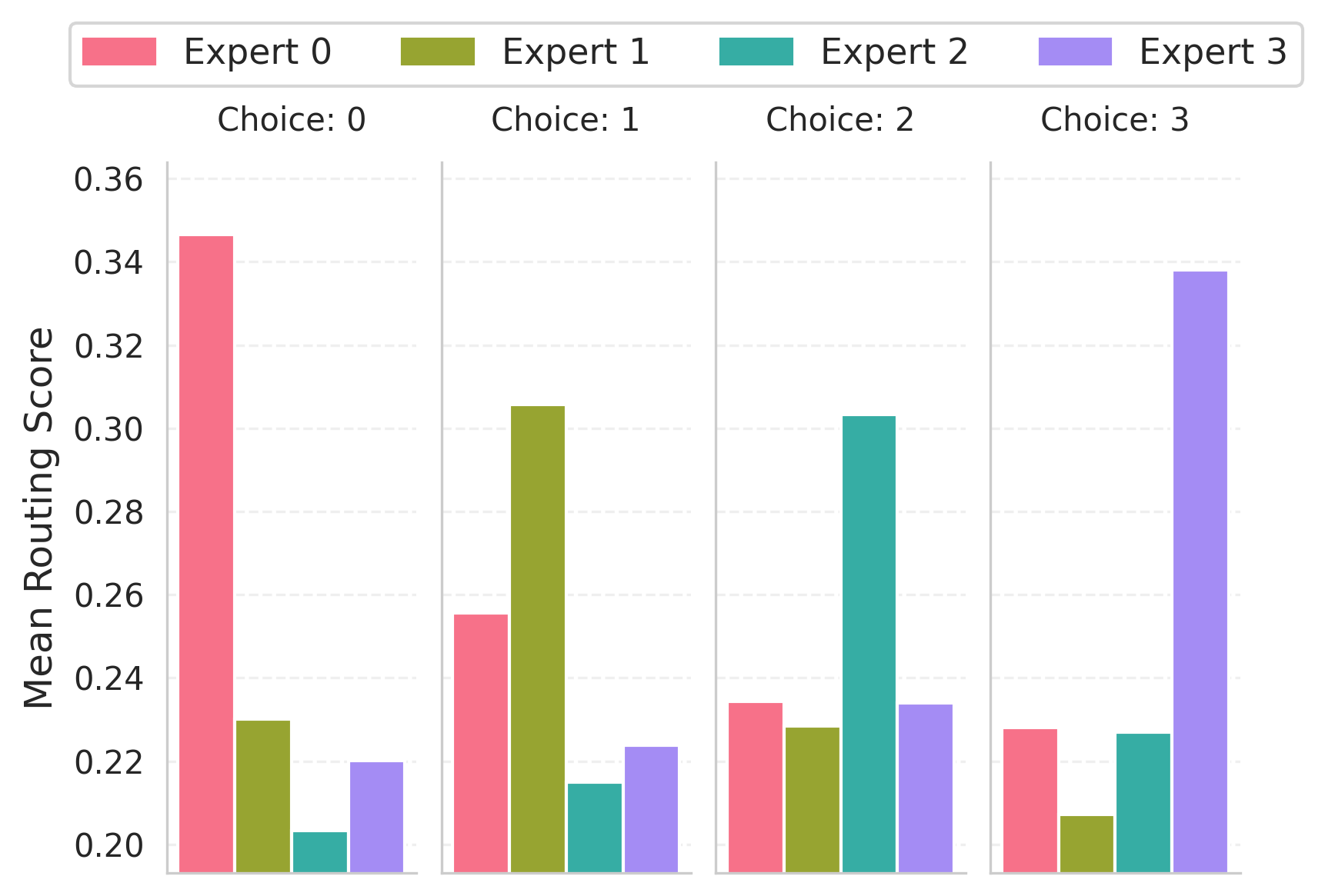}
        \subcaption{Routing scores.}
        \label{fig:moe_expert_routing_score}
    \end{minipage}
    \caption{Analysis of DMMM expert features. (a) Distinct features across experts. (b) Expert routing scores. Both demonstrate the effectiveness of the disentangled design.}
    \label{fig:moe_routing}
    % \vspace{-10pt}
% \end{wrapfigure}
\end{figure}

\paragraph{Expert-Wise Effectiveness.} 
% To validate DMMM's expert-wise effectiveness to model complex spatio-temporal patterns, we visualized first-layer features in STM3 using t-SNE \cite{van2008visualizing}. Figure~\ref{fig:moe_visulization} shows distinct feature clusters for each expert, confirming successful pattern disentanglement through our contrastive learning. Figure~\ref{fig:moe_expert_routing_score} further demonstrates the gating network's discriminative capability, with significantly varied assignment scores across experts for samples routed to the corresponding expert, indicating confident expert selection.
To validate DMMM’s expert-wise effectiveness to model complex spatio-temporal patterns, we visualized STM3’s first-layer features using t-SNE~\cite{van2008visualizing}. Figure~\ref{fig:moe_visulization} shows distinct feature clusters for each expert, confirming effective pattern disentanglement. Figure~\ref{fig:moe_expert_routing_score} further illustrates the gating network's discriminative capability, with significantly varied assignment scores across experts for samples routed to the corresponding expert, indicating confident expert selection.

\begin{figure}[t]
% \begin{wrapfigure}{r}{0.49\textwidth}
%     \vspace{-20pt}
    \centering
    \begin{minipage}[c]{0.24\textwidth}
        \centering
        \includegraphics[height=4.1cm]{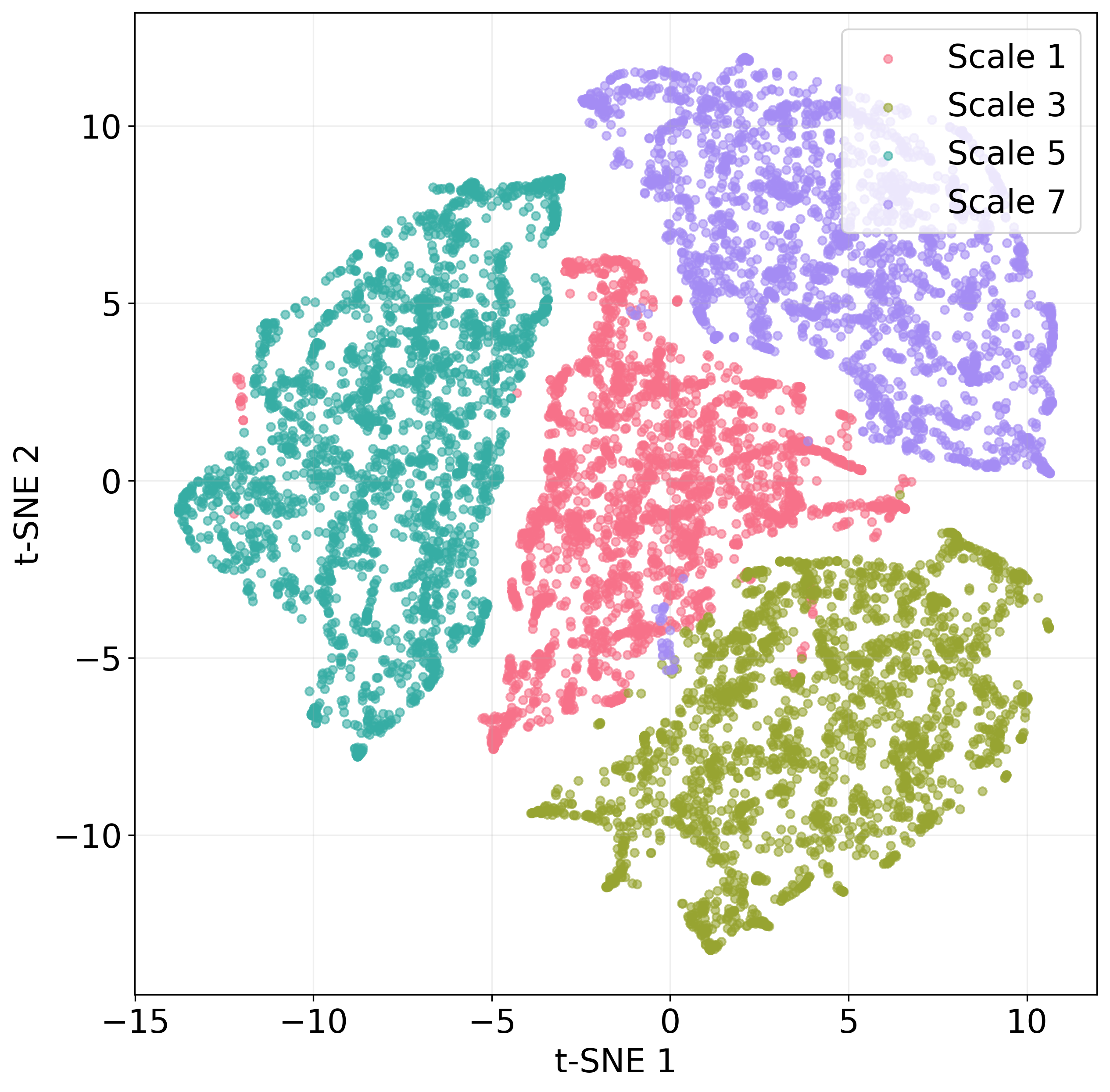}
        \subcaption{Scale-wise features.}
        \label{fig:moe_tsne_scale_wise}
    \end{minipage}
    \hfill
    \begin{minipage}[c]{0.23\textwidth}
        % \vspace{14pt}
        \centering
        \includegraphics[height=4.1cm]{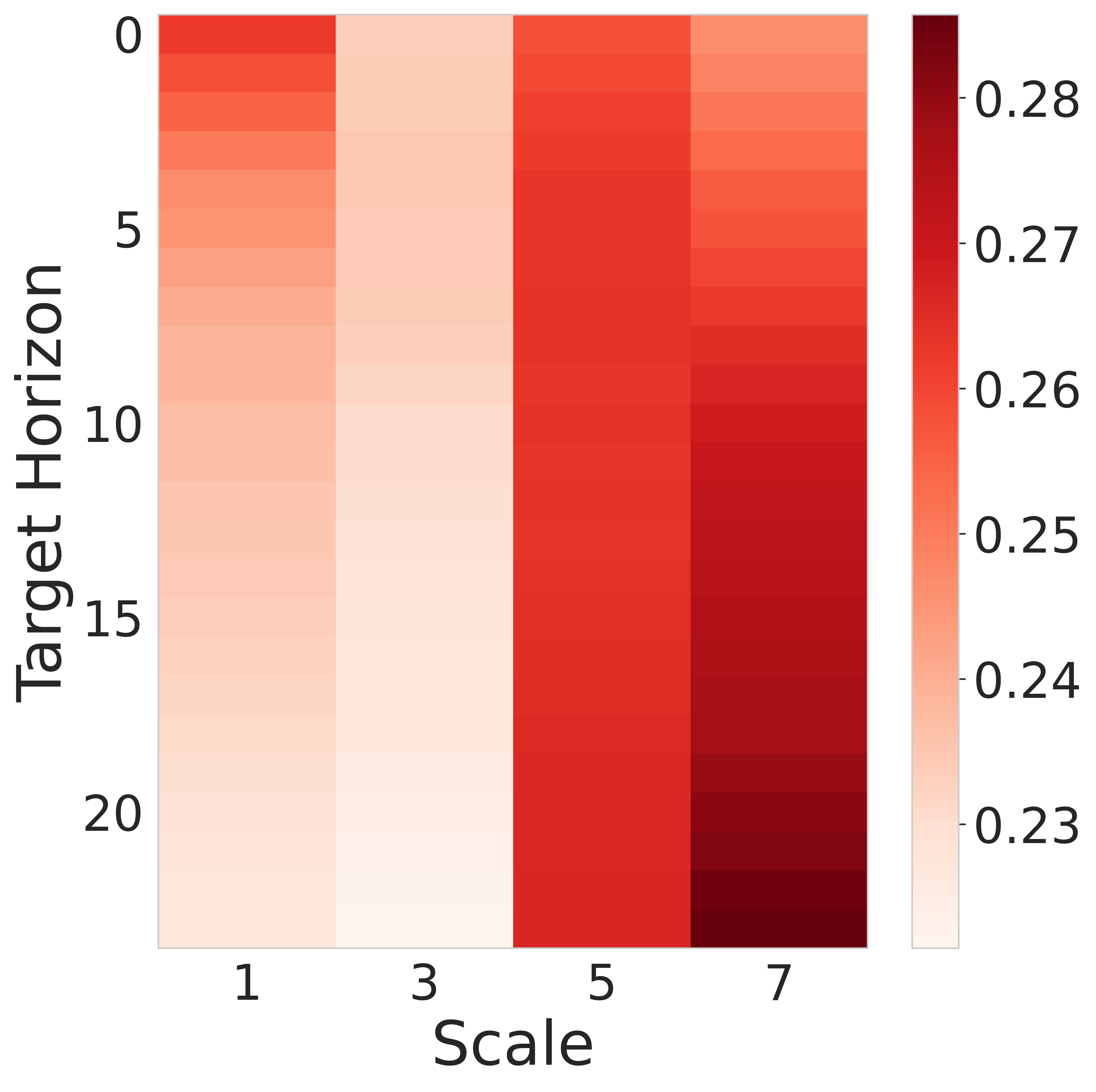}
        \subcaption{Scale weights.}
        \label{fig:scale_weight}
    \end{minipage}
    \caption{Analysis of STM3's multiscale feature extraction. (a) Distinct features across scales, demonstrating effective disentanglement. (b) Contribution of disentangled multiscale patterns to predictions at different horizons.}
    \label{fig:scale-wise}
    \vspace{-10pt}
% \end{wrapfigure}
\end{figure}

\paragraph{Multiscale Feature Extraction.} 
To evaluate STM3's multiscale extraction capability, we analyzed features from the first DMMM expert by aggregating and visualizing scale-indexed features via t-SNE (Figure~\ref{fig:moe_tsne_scale_wise}). The results confirm that experts extract scale-specific features, validating our contrastive learning approach for scale-wise disentanglement. Moreover, we modified STM3 by inserting a learnable weighting parameter $\Gamma \in \mathbb{R}^{\tau \times Q}$ in the output layer, performing softmax-weighted scale fusion over scales for final predictions. Figure~\ref{fig:scale_weight} reveals an adaptive scale selection mechanism where larger scales become increasingly important for longer prediction horizons. This demonstrates STM3's capability to differentially process multiscale features based on temporal context.

\paragraph{Routing Strategy.}

\begin{figure}[t]
    \centering
    \begin{minipage}[c]{0.48\columnwidth}
        \centering
        \includegraphics[height=4.1cm]{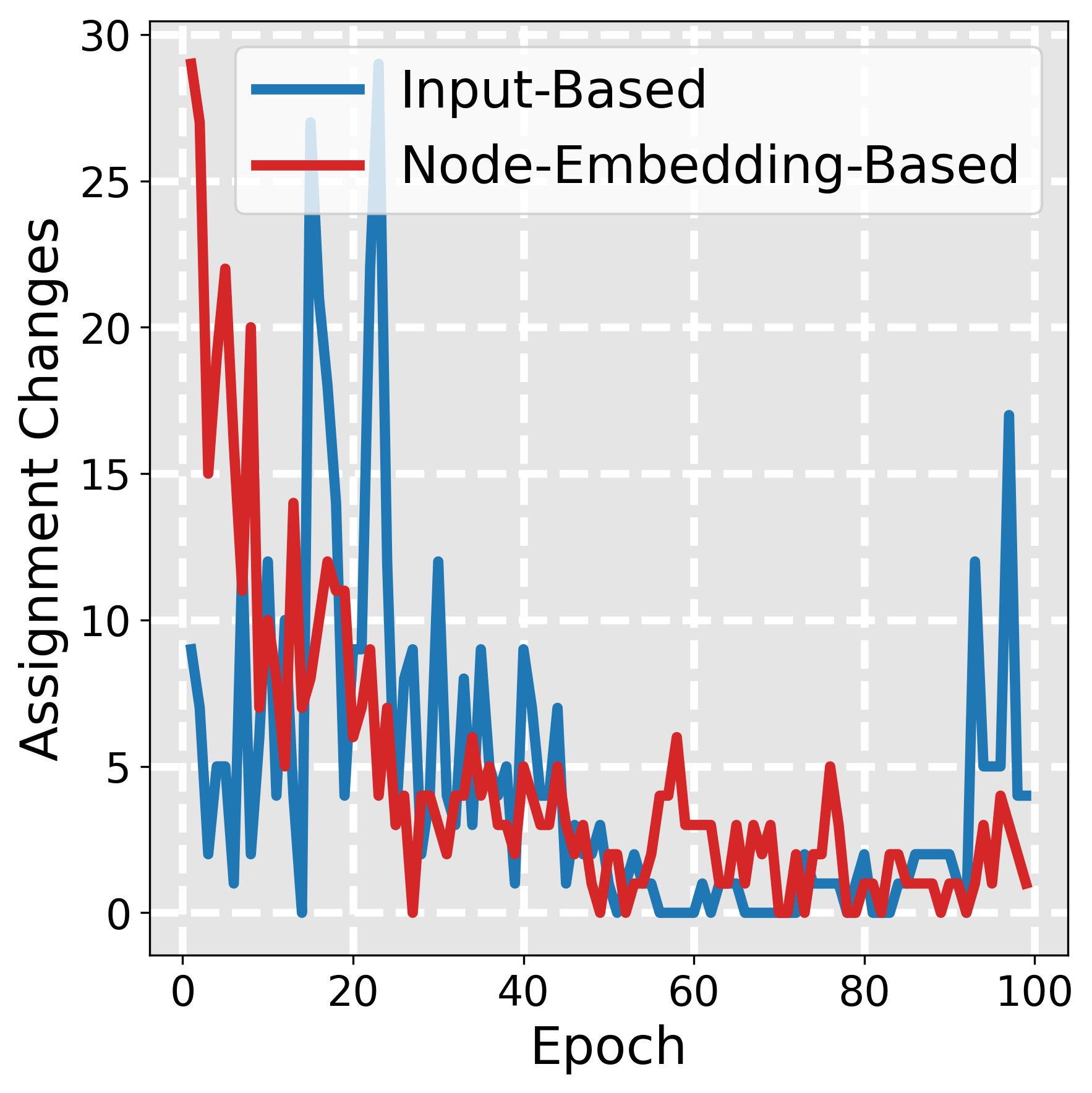}
        \subcaption{Expert assignment.}
        \label{fig:moe_expert_convergence}
    \end{minipage}
    \hfill
    \begin{minipage}[c]{0.48\columnwidth}
        \centering
        \includegraphics[height=4.1cm]{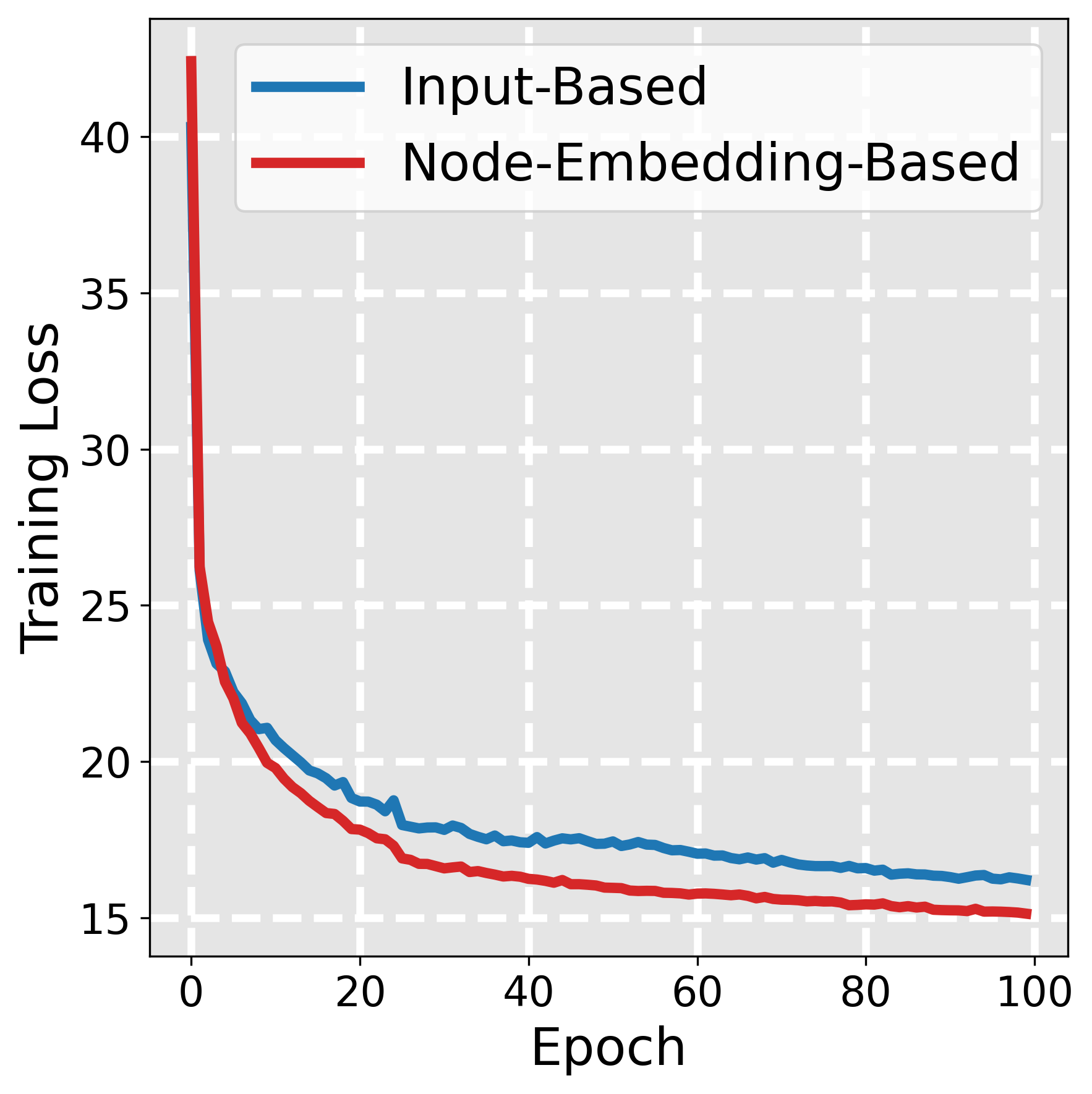}
        \subcaption{Loss.}
        \label{fig:moe_loss_convergence}
    \end{minipage}
    
    \caption{Comparison of routing strategies. (a) shows that the assignment of our Node-Embedding-Based routing strategy is more stable than the traditional Input-Based one during training. (b) shows that this stability contributes to better training performance.}
    \vspace{-10pt}
    \label{fig:moe_convergence}
\end{figure}

To evaluate our node-embedding-based routing against traditional input-based routing, we compared their convergence through expert assignment stability and loss trajectories. For the first test sample's MoE layer, we tracked epoch-wise expert assignments and computed inter-epoch "assignment changes" (nodes with altered allocations). Figure~\ref{fig:moe_expert_convergence} shows our strategy achieves rapid convergence with minimal assignment fluctuations, while traditional routing exhibits persistent oscillations that persist even in later epochs. This observation confirms the better routing smoothness of our routing strategy. Similarly, Figure~\ref{fig:moe_loss_convergence} demonstrates our approach's superior loss convergence with faster convergence values and lower final, confirming its performance advantage.

\section{Conclusion}\label{sec:conclusion}
In this work, we propose STM3, a comprehensive solution for long-term spatio-temporal prediction grounded in the fundamental insight that inter-node spatial correlations are intrinsically driven by the similarity of multiscale temporal patterns. To realize this, STM3 features a tripartite innovation with Multiscale Mamba, Adaptive Graph Causal Convolution Network, and Disentanglement MoE. Theoretical analysis confirms that our stable routing strategy and causal contrastive learning ensure superior routing smoothness and effective pattern disentanglement. Extensive experiments validate this synergy, achieving state-of-the-art performance. Notably, on the PEMSD8 dataset, STM3 surpasses the runner-up by 7.1\% in MAE, 8.5\% in RMSE, and 15.9\% in MAPE. This work advances MoE research in spatio-temporal domains, offering valuable insights into future developments.

% \begin{acks}
% xxx
% \end{acks}

\bibliographystyle{ACM-Reference-Format}
\bibliography{sample-base}

\appendix

\section{More Related Work}\label{appendix:more_related_work}

\paragraph{State Space Models.}

SSMs have demonstrated exceptional capability in modeling sequential dependencies via state space. The structured state-space sequence model (S4)~\cite{gu2021efficiently} pioneered efficient long-range dependency modeling with linear complexity. Subsequent advancements led to variants such as S5~\cite{smith2023simplified}, H3~\cite{dao2023hungry}, and GSS~\cite{mehta2023long}. Mamba~\cite{gumamba} marked a breakthrough by introducing data-dependent SSM parameters and parallel scan selection (S6), surpassing quadratic-complexity transformers in processing long sequences while maintaining linear computational efficiency. In the domain of spatio-temporal and time-series modeling, Mamba has emerged as a powerful component~\cite{cai2024mambats, xu2025sst, alkilane2024mixmamba, li2024stg, choi2024spot, He_Ji_Lei_2025}. For instance, SST~\cite{xu2025sst} employs hybrid Mamba-Transformer experts to model long- and short-range time series patterns, adaptively balancing the contributions of the two experts via a long-short router to integrate global trends and local variations. MixMamba~\cite{alkilane2024mixmamba} utilizes multiple Mamba layers as experts, incorporating MoE mechanism to effectively model heterogeneous and non-stationary time series data. STG-Mamba~\cite{li2024stg} combines GNNs with selective SSMs, dynamically updating adjacency matrices through Kalman filter-based graph neural networks. DST-Mamba~\cite{He_Ji_Lei_2025} employed predefined graph structures for spatial relationships. In our work, we further identify and exploit the connection between different scales, deriving a novel Multiscale Mamba architecture for spatio-temporal dependency modeling.

\paragraph{Mixture-of-Experts.}

The MoE is a classical neural network architecture~\cite{jacobs1991adaptive} designed to increase model capacity with minimal computational overhead. Recently, MoE networks have become integral components in deep neural networks, notably in large language models, achieving remarkable success~\cite{du2022glam, fedus2022switch, lepikhin2020gshard, zoph2022designing, jiang2024mixtral, dai2024deepseekmoe}. By employing a gating network to route different data subsets to a small group of expert networks with identical architectures, MoE enables specialized experts to effectively model distinct data patterns, thus enhancing overall model capacity. Recent theoretical studies have further analyzed the superior performance of MoE structures, elucidating their inherent advantages~\cite{chen2022towards, nguyen2023statistical, nguyen2024least, li2024theory}. In the time-series modeling domain, several studies have leveraged MoE's capabilities to handle complex data distributions~\cite{shi2024time, xu2025sst, alkilane2024mixmamba}. In this work, we propose a novel routing strategy based on node embeddings, which enforces each routed expert to capture data distributions at the granularity of spatial nodes. By utilizing our proposed Multiscale Mamba as experts, our model demonstrates strong fitting capabilities for complex multiscale patterns.

\section{Proof}

\subsection{Proof of Lemma 1}\label{appendix:proof:gating}

% \begin{proof}
We proceed with three steps:

1. \textbf{Pairwise Bound}: 
Each pairwise disagreement is bounded as:
\begin{equation}
p_{i,j} \leq \rho K^2 \|\mathbf{h}_i - \mathbf{h}_j\|_\infty
\end{equation}

Given random variable $\{r_k\}^K_{k=1}$, let $k_1 = \underset{m \in [K]}{\mathrm{argmax}}\{h_{i,k} + r_k\}$ and $k_2 = \underset{k \in [K]}{\mathrm{argmax}}\{{h}_{j,k} + r_k\}$, then we have that
\begin{equation}
h_{i,k_1} + r_{k_1} > h_{i,k_2} + r_{k_2}, \quad  {h}_{j,k_1} + r_{k_1} > {h}_{j,k_2} + r_{k_2},
\end{equation}
which implies that
\begin{equation}\label{eq:inequality_3}
{h}_{j,k_2} - {h}_{j,k_1} > r_{k_1} - r_{k_2} > h_{i,k_2} - h_{i,k_1}.
\end{equation}

Define $C(k_1, k_2) = \frac{{h}_{j,k_2} - {h}_{j,k_1} + h_{i,k_2} - h_{i,k_1}}{2}$, then (\ref{eq:inequality_3}) implies that:
\begin{equation}\label{eq:c2_bound}
|r_{k_1} - r_{k_2} - C(k_1, k_2)| \leq \frac{|{h}_{j, k_2} - {h}_{j, k_1} - h_{i, k_2} + h_{i, k_1}|}{2} \leq \|\mathbf{h}_i - \mathbf{h}_j\|_\infty
\end{equation}

For any $i,j \in [M]$, the pairwise disagreement probability is bounded as:
% \begin{align*}
% p_{i,j}
% & = \mathbb{P}\left(\underset{k\in[K]}{\mathrm{argmax}}\{h_{i,k} + r_k\} \neq \underset{k\in[K]}{\mathrm{argmax}}\{h_{j,k} + r_k\}\right) \\
% & \leq \mathbb{P}\left( \exists k_1 \neq k_2 \in [K], \text{s.t.} | r_{k_1} - r_{k_2} - C(k_1, k_2) | \leq \|\mathbf{h}_i - \mathbf{h}_j\|_\infty  \right)  \\
% &\leq  \sum_{k_1 < k_2}\mathbb{P} \left( | r_{k_1} - r_{k_2} - C(k_1, k_2) | \leq \|\mathbf{h}_i - \mathbf{h}_j\|_\infty \right)   \\
% &=  \sum_{k_1 < k_2} \mathbb{E}\left[ r_{k_2} + C(k_1, k_2) - \|\mathbf{h}_i - \mathbf{h}_j\|_\infty \leq r_{k_1} \leq r_{k_2} + C(k_1, k_2) - |\|\mathbf{h}_i - \mathbf{h}_j\|_\infty \Big| r_{k_2} \right]  \\
% &\leq \sum_{k_1 < k_2} 2\rho \|\mathbf{h}_i - \mathbf{h}_j\|_\infty \\
% &\leq \rho K^2 \|\mathbf{h}_i - \mathbf{h}_j\|_\infty,
% \end{align*}
% 换行更改版
\begin{align*}
p_{i,j}
&= \mathbb{P}\left(
\underset{k\in[K]}{\mathrm{argmax}} \{h_{i,k} + r_k\} 
\neq 
\underset{k\in[K]}{\mathrm{argmax}} \{h_{j,k} + r_k\}
\right) \\
&\leq \mathbb{P}\left(
\exists\, k_1 \neq k_2 \in [K],\ \text{s.t. } 
\left| r_{k_1} - r_{k_2} - C(k_1, k_2) \right| 
\leq \|\mathbf{h}_i - \mathbf{h}_j\|_\infty 
\right) \\
&\leq \sum_{k_1 < k_2} \mathbb{P}\left(
\left| r_{k_1} - r_{k_2} - C(k_1, k_2) \right| 
\leq \|\mathbf{h}_i - \mathbf{h}_j\|_\infty 
\right) \\
&= \sum_{k_1 < k_2} \mathbb{E}\Big[
r_{k_2} + C(k_1, k_2) - \|\mathbf{h}_i - \mathbf{h}_j\|_\infty \leq r_{k_1} \\
&\hspace{3.9em} \leq r_{k_2} + C(k_1, k_2) + \|\mathbf{h}_i - \mathbf{h}_j\|_\infty \Big] \\
&\leq \sum_{k_1 < k_2} 2\rho \|\mathbf{h}_i - \mathbf{h}_j\|_\infty \\
&\leq \rho K^2 \|\mathbf{h}_i - \mathbf{h}_j\|_\infty,
\end{align*}

2. \textbf{Upper Bound}:
\begin{align*}
\mathbf{P} 
&= \mathbb{P}\left(\exists i \neq j \in [M], s.t.  \underset{k\in[K]}{\mathrm{argmax}}\{h_{i,k} + r_{k}\} \neq \underset{k\in[K]}{\mathrm{argmax}}\{h_{j,k} + r_{k}\}\right) \\
&\leq \sum_{i < j} \mathbb{P} \left( \underset{k\in[K]}{\mathrm{argmax}}\{h_{i,k} + r_k\} \neq \underset{k\in[K]}{\mathrm{argmax}}\{h_{j,k} + r_k\} \right) \\
&\leq \sum_{i < j} p_{i,j} \\
&\leq \rho K^2 \sum_{i < j} \|\mathbf{h}_i - \mathbf{h}_j\|_\infty,
\end{align*}

3. \textbf{Lower Bound}: When all $\mathbf{h}_i$ are identical, $\|\mathbf{h}_i - \mathbf{h}_j\|_\infty = 0$ and the probability equals 0. This condition is met for our routing strategy, since the gating output $\mathbf{h}_i$ of every sample is generated by the same node embedding $E_{A}$. Obviously, for the traditional input-based routing strategy, such a condition cannot be achieved, since it means all input time series are the same.

% \end{proof}

\subsection{Proof of Lemma 2}\label{appendix:proof:disentangle}

% \begin{proof}
Consider the contrastive loss
\[
\mathcal{L}_{\text{contr}} = -\log\frac{\exp(s(x_i, x_j))}{\sum_{x \in \mathcal{X}} \exp(s(x_i,x))},
\]
where $x_i = \phi(I_i)$ and $x_j = \phi(I_j)$ are features extracted from two inputs $I_i$ and $I_j$ belonging to the same orbit under the subgroup ${D} \subset {G}$. The loss encourages the similarity between $x_i$ and $x_j$ by maximizing the similarity $s(x_i, x_j)$ while simultaneously minimizing the similarity to all other features in the denominator. Since $I_j = d \cdot I_i$ for some $d \in {D}$, this enforces that the expert $\phi$ must produce features that are equivariant under ${D}$-actions; otherwise, $s(x_i, x_j)$ would be suboptimal, increasing the loss. Therefore, minimizing $\mathcal{L}_{\text{contr}}$ enforces that $\phi(d \cdot I) = d \cdot \phi(I)$ for all $d \in {D}$, leading to ${D}$-equivariance in the representation space ${X}$.

Simultaneously, for negative samples $x_k = \phi(I_k)$ where $I_k$ belongs to a different orbit under ${D}$—specifically, $I_k = c \cdot I_i$ for some $c \in {G}/{D}$—the loss penalizes the similarity $\exp(x_i^\top x_k)$. This encourages the model to minimize any alignment between $x_i$ and features from other orbits, thus pushing the representation $\phi$ to become invariant under the action of ${G}/{D}$. In other words, $\phi(c \cdot I) = \phi(I)$ must hold for all $c \in {G}/{D}$ to effectively suppress cross-orbit similarity in the contrastive objective.

Combining the above, we conclude that the learned feature space $\mathcal{X}$ admits a decomposition of the form
\[
\mathcal{X} = \mathcal{X}_{{G}/{D}} \times \mathcal{X}_{{D}},
\]
where $\mathcal{X}_{{G}/{D}}$ is invariant to ${D}$-actions and sensitive to ${G}/{D}$, while $\mathcal{X}_{{D}}$ is equivariant to ${D}$-actions and invariant to ${G}/{D}$. To verify this decomposition formally, let $g = (c, d) \in {G} = ({G}/{D}) \times {D}$. Then for any $x = \phi(I)$,
\[
g \cdot x = \phi(g \cdot I) = \phi(c \cdot (d \cdot I)) = \phi(d \cdot I) = d \cdot \phi(I) = d \cdot x,
\]
where the third equality follows from ${G}/{D}$-invariance and the fourth from ${D}$-equivariance. This confirms that the group action affects only the ${D}$-equivariant component $\mathcal{X}_{{D}}$, while the ${G}/{D}$-invariant component remains unaffected. Hence, the contrastive loss induces a disentangled representation aligned with the product structure of the group ${G}$, satisfying the desired properties of equivariance and decomposability.

% \end{proof}

\section{Implementation Details}\label{appendix:implement_detail}

\paragraph{Data.} 
Several large-scale real-world experiments were selected to evaluate the effectiveness of STM3. The METR-LA dataset comprises traffic statistics collected from the road network of Los Angeles County, reflecting daily spatio-temporal patterns~\cite{li2018diffusion}. Meanwhile, the PEMSD4 and PEMSD8 datasets are part of the Caltrans Performance Measurement System~\cite{chen2001freeway}. The KnowAir dataset records the PM2.5 of several regions in China~\cite{wang2020pm2}. The real-world wireless traffic dataset collected by Telecom Italia in Milan~\cite{barlacchi2015multi}. The dataset comprises three types of cellular services: SMS, voice calls, and Internet usage. For our experiments, we randomly select 300 cells from a total of 10,000 across the city. The NREL dataset registered solar power output by photovoltaic power plants in Alabama~\cite{wu2021inductive}. These datasets are widely adopted in prior research. Detailed statistical descriptions are provided in Table \ref{tab:dataset_info}.

\begin{table*}[t]
% \small
\centering
\caption{Statistical Information of Experimental Dataset}
\label{tab:dataset_info}
\begin{tabular}{lllll}
\toprule
Dataset & Data Record & Node & Sample Rate & Time Range \\
\midrule
METR-LA & transport traffic speed & 207 & 5-minute & 2012.03 - 2012.06 \\
PEMS04 & transport traffic volume & 307 & 5-minute & 2018.01 - 2018.02 \\
PEMS08 & transport traffic volume & 170 & 5-minute & 2016.07 - 2016.08 \\ 
KnowAir & air quality & 184 & 3-hour & 2015.01 - 2018.12 \\ 
Milan & mobile traffic volume & 300 & 1-hour & 2013.11 - 2014.01 \\ 
NREL & solar power output & 137 & 10-minute & 2006.01 - 2006.03 \\ 
ETTh1 & electricity & 7 & 1-hour & 2016.07 - 2018.06 \\ 
Electricity & electricity & 321 & 1-hour & 2016.07 - 2019.07 \\ 
\bottomrule
\end{tabular}
\end{table*}

\paragraph{Baselines.}
Our STM3 model is evaluated on a set of advanced baselines as follows:
\begin{itemize}[itemsep=0em, topsep=0em, leftmargin=1em]
    \item \textbf{DST-Mamba}\cite{He_Ji_Lei_2025} decomposes the temporal pattern into seasonal and trend parts, using bi-directional Mamba and a graph method for temporal and spatial dependency modeling.
    
    \item \textbf{SST}\cite{xu2025sst} using hybrid Mamba and Transformer experts to model temporal dependency.
    
    \item \textbf{MixMamba}\cite{alkilane2024mixmamba} uses several Mamba layers as experts, incorporating an MoE mechanism to model the heterogeneous and non-stationary time series data.
    
    \item \textbf{STGMamba}\cite{li2024stg} uses Kalman filtering GNN to model spatial dependency and uses Mamba to model temporal dependency.
    
    \item \textbf{DGraFormer}\cite{yan2025dgraformer} uses dynamic graph learning to model spatial dependency and uses Transformer to model temporal dependency.
    
    \item \textbf{FilterTS}\cite{Wang_Liu_Duan_Wang_2025} converts the time-domain series into the frequency domain and uses filters to model temporal dependency.
    
    \item \textbf{AMD}\cite{Hu_Liu_Zhu_Cheng_Dai_2025} decomposes multi-scale temporal information and uses MLP to model temporal dependency.
    
    \item \textbf{TimeMixer++}\cite{wang2025timemixer++} disentangles seasonality and trend via dual-axis attention and uses MLP to model temporal dependency.
    
    \item \textbf{iTransformer}\cite{liu2024itransformer} inverts the temporal and feature dimension, and uses Transformer to model temporal dependency.
    
    \item \textbf{STGM}\cite{lablack2023spatio} uses similarity estimator to model spatial dependency and uses Transformer to model temporal dependency.
    
    \item \textbf{PatchTST}\cite{nie2023a} converts time-series into patches and uses Transformer to model temporal dependency.
    
    \item \textbf{PromptST}\cite{zhang2023promptst} dynamically generates task-specific prompts to capture spatio-temporal dependency.
    
    \item \textbf{Informer}\cite{Zhou_Zhang_Peng_Zhang_Li_Xiong_Zhang_2021} uses encoder-decoder type Transformer to model temporal dependency.
\end{itemize}

\textbf{Implementation Details.} To ensure a fair and reproducible comparison, we strictly adhered to the optimal hyperparameters provided in the official scripts and papers for all baseline models. Dataset-specific configurations reported by the original authors were applied where available. We made minimal modifications only when necessary, with the explicit goal of \textbf{facilitating the best possible performance for the baselines}:
\begin{itemize}[itemsep=0em, topsep=0em, leftmargin=1em]
    \item \textbf{Fairness (Sufficient Training):} We standardized the training duration by increasing \texttt{train\_epochs} to $100$ for baselines including FilterTS, Informer, iTransformer, MixMamba, AMD, and SST. This ensures these models strictly avoid under-fitting and are fully converged.
    \item \textbf{Stability (Convergence Guarantee):} We adjusted hyperparameters solely to rescue models from bugs or non-convergence issues. Specifically, we modified the patch stride for DGraFormer on the Milan dataset, and adjusted the learning rates for iTransformer (on PEMSD4/8) and AMD. These changes ensure the baselines operate stably and competitively.
\end{itemize}
Detailed configurations are provided in Table~\ref{tab:baseline_configs}.

\begin{table*}[h]
\centering
\caption{Detailed Hyperparameter Configurations and Modifications for Baselines.}
\label{tab:baseline_configs}
\setlength{\tabcolsep}{4pt} 
\fontsize{8pt}{9pt}\selectfont 
\renewcommand{\arraystretch}{1.1} % 为了提升密集参数的可读性，稍微增加了行间距

% 调整了列宽：中间列用于显示较长的原始参数，右侧列用于显示修改说明
\begin{tabular}{l p{9.5cm} p{5.5cm}}
\toprule
\textbf{Model} & \textbf{Original Configuration (Key Hyperparameters)} & \textbf{Modifications \& Notes} \\
\midrule
\textbf{DGraFormer} & d\_model=16, n\_heads=4, e\_layers=1, d\_ff=128, \text{patch}=8, \text{stride}=8, \text{dropout}=0.05. & For \textit{Milan} dataset (input len=12), changed to $\text{patch}=1, \text{stride}=1$ to prevent bugs. \\
\midrule
\textbf{DST-Mamba} & d\_model=512, e\_layers=2, d\_state=32, \text{rank}=32, \text{ds\_layers}=3. Specific settings for MERT-LA (e\_layers=3) and PEMS (e\_layers=2). & None \\
\midrule
\textbf{FilterTS} & d\_model=512, e\_layers=2, \text{factor}=1, \text{quantile}=0.9, \text{top\_K\_static\_freqs}=10, \text{bandwidth}=1. & Increased \texttt{train\_epochs} to 100 for fairness. \\
\midrule
\textbf{Informer} & d\_model=512, n\_heads=8, e\_layers=2, d\_ff=2048, \text{factor}=5. & Increased \texttt{train\_epochs} to 100 for fairness. \\
\midrule
\textbf{iTransformer} & Default: d\_model=512, n\_heads=8, e\_layers=2. Specific settings for ECL (e\_layers=3), PEMSD4 (e\_layers=4, d\_model=1024), PEMSD8 (e\_layers=4). & 1. Increased \texttt{train\_epochs} to 100 for fairness. \newline 2. Reset learning rate to default ($10^{-4}$) for PEMSD4/8 to fix convergence issues. \\
\midrule
\textbf{MixMamba} & Default: d\_model=512, \text{num\_experts}=16, d\_state=16, expand=3. Specific settings for ECL (d\_model=32, num\_experts=8, d\_state = 8, expand = 2), ETTh1 (d\_model=64, num\_experts=2, d\_state=16, expand = 2). & Increased \texttt{train\_epochs} to 100 for fairness. \\
\midrule
\textbf{AMD} & Default: \text{patch}=16, \text{mix\_layers}=3, \text{scale}=2, norm=True, layernorm=True. Specific settings for ECL(layernorm=False) provided. & 1. Increased \texttt{train\_epochs} to 100 for fairness. \newline 2. Adjusted learning rate to $0.003$ to fix convergence issues. \\
\midrule
\textbf{PatchTST} & Default: n\_heads=16, d\_model=128, d\_ff=256, \text{dropout}=0.2, fc\_dropout=0.2. Specific settings for ETTh1 (n\_heads=4, d\_model=16, d\_ff=128, dropout=0.3, fc\_dropout=0.3). & None. \\
\midrule
\textbf{PromptST} & \text{hid\_dim}=32, \text{ts\_depth\_spa}=2, \text{ts\_depth\_tem}=2. & None. \\
\midrule
\textbf{SST} & Default: n\_heads=4, d\_model=16, d\_ff=128, \text{local\_ws}=7. Specific settings for ECL (n\_heads=16, d\_model=128, d\_ff=256, dropout=0.2, fc\_dropout=0.2). & Increased \texttt{train\_epochs} to 100 for fairness. \\
\midrule
\textbf{STGM} & \text{hidden\_dim}=64, \text{num\_blocks}=2. & None. \\
\midrule
\textbf{STGMamba} & K=3, n\_layer=4. & None. \\
\midrule
\textbf{TimeMixer++} & n\_layers=3, d\_model=16, \text{top\_k}=5, \text{dropout}=0.1. & None. \\
\bottomrule
\end{tabular}
\end{table*}

\paragraph{Model Settings.}
The STM3 with default settings: the number of layers $L=2$, the dimension of the feature of each spatio-temporal step $d=4$, the dimension of the feature of each spatial node $d_e=12$, the number of routed experts $K=4$, the number of scales $Q=4$ with the scale list $S=[1, 3, 5, 7]$, the contrastive learning loss weight $\lambda = 0.1$, and the causal similarity weights $\gamma_1 = \gamma_2=0.1$.

\section{Additional Experiments}

\subsection{Full Comparison Experiment}\label{appendix:full_compare}

We provide the full experimental results in Table~\ref{tab:detailed_results}.

\begin{table*}[htbp]
  \centering
  \caption{Detailed experimental results on all benchmarks and in-out settings.}
  \label{tab:detailed_results}
  \resizebox{\textwidth}{!}{%
    \begin{tabular}{llccccccccccccccc}
    \toprule
    \textbf{Benchmark} & \textbf{In-Out} & \textbf{Metric} & \textbf{STM3} & \textbf{DST-Mamba} & \textbf{SST} & \textbf{MixMamba} & \textbf{STGMamba} & \textbf{DGraFormer} & \textbf{FilterTS} & \textbf{AMD} & \textbf{TimeMixer++} & \textbf{iTransformer} & \textbf{STGM} & \textbf{PatchTST} & \textbf{PromptST} & \textbf{Informer} \\
    \midrule
    \multirow{15}{*}{METR\_LA} & \multirow{3}{*}{96$\to$96} & MAE & \textbf{\textcolor{red}{9.5837}} & 12.2376 & 12.2008 & 11.7474 & 11.7356 & 11.2048 & 12.6240 & 12.4063 & 13.7090 & 12.3821 & 11.2826 & 12.6610 & \underline{\textcolor{blue}{10.6535}} & 13.2870 \\
     &  & RMSE & \textbf{\textcolor{red}{15.5923}} & 17.2296 & 18.1973 & 17.4471 & 17.3303 & 18.7093 & 18.1864 & 19.0260 & 19.8236 & 17.2701 & \underline{\textcolor{blue}{16.1209}} & 18.0149 & 17.9203 & 17.9001 \\
     &  & MAPE & \textbf{\textcolor{red}{0.1772}} & 0.2449 & 0.2777 & 0.2436 & \underline{\textcolor{blue}{0.1875}} & 0.2663 & 0.2689 & 0.2787 & 0.2545 & 0.2538 & 0.2385 & 0.2800 & 0.2579 & 0.2220 \\
    \cmidrule{2-17}
     & \multirow{3}{*}{96$\to$192} & MAE & \textbf{\textcolor{red}{10.8418}} & 13.8839 & 14.0644 & 12.8499 & 13.1196 & \underline{\textcolor{blue}{12.3956}} & 13.7923 & 13.7053 & 14.2397 & 13.5994 & 12.4501 & 14.0348 & 12.7412 & 14.5077 \\
     &  & RMSE & \textbf{\textcolor{red}{17.1490}} & 18.5261 & 19.9018 & 18.7197 & 18.2372 & 19.9584 & 18.9567 & 20.2441 & 20.0859 & 18.4049 & \underline{\textcolor{blue}{17.4241}} & 19.3355 & 19.8398 & 18.7967 \\
     &  & MAPE & \textbf{\textcolor{red}{0.1770}} & 0.2593 & 0.2906 & 0.2570 & \underline{\textcolor{blue}{0.2102}} & 0.2605 & 0.2684 & 0.2894 & 0.2598 & 0.2622 & 0.2423 & 0.2902 & 0.2661 & 0.2503 \\
    \cmidrule{2-17}
     & \multirow{3}{*}{96$\to$336} & MAE & \textbf{\textcolor{red}{11.2723}} & 13.8600 & 14.1818 & 12.9408 & 13.2861 & \underline{\textcolor{blue}{12.3367}} & 13.9140 & 13.7869 & 14.7684 & 13.7656 & 12.6418 & 14.2149 & 13.4050 & 14.8667 \\
     &  & RMSE & \textbf{\textcolor{red}{17.4188}} & 18.5278 & 19.8195 & 18.8751 & 18.5075 & 19.7686 & 19.0455 & 20.1995 & 20.2792 & 18.6224 & \underline{\textcolor{blue}{17.6236}} & 19.5119 & 20.5269 & 19.3373 \\
     &  & MAPE & \textbf{\textcolor{red}{0.1662}} & 0.2408 & 0.2799 & 0.2430 & \underline{\textcolor{blue}{0.2006}} & 0.2343 & 0.2544 & 0.2718 & 0.2692 & 0.2473 & 0.2189 & 0.2757 & 0.2597 & 0.2270 \\
    \cmidrule{2-17}
     & \multirow{3}{*}{96$\to$720} & MAE & \textbf{\textcolor{red}{12.2057}} & 15.4833 & 15.5984 & 14.3447 & 15.6095 & 13.8990 & 15.6588 & 15.6474 & 14.0795 & 15.8224 & \underline{\textcolor{blue}{13.5989}} & 15.7836 & 13.7145 & 15.4446 \\
     &  & RMSE & \textbf{\textcolor{red}{18.7180}} & 20.2465 & 21.5643 & 20.3718 & 19.8006 & 22.9704 & 20.7050 & 22.6300 & 20.6308 & 20.5581 & \underline{\textcolor{blue}{18.8801}} & 21.1229 & 21.0475 & 19.7240 \\
     &  & MAPE & \textbf{\textcolor{red}{0.1783}} & 0.2576 & 0.2886 & 0.2531 & 0.2610 & 0.2574 & 0.2675 & 0.2923 & 0.2413 & 0.2696 & \underline{\textcolor{blue}{0.2266}} & 0.2890 & 0.2336 & 0.2418 \\
    \cmidrule{2-17}
     & \multirow{3}{*}{Avg} & MAE & \textbf{\textcolor{red}{10.9759}} & 13.8662 & 14.0114 & 12.9707 & 13.4377 & \underline{\textcolor{blue}{12.4590}} & 13.9973 & 13.8865 & 14.1992 & 13.8924 & 12.4934 & 14.1736 & 12.6286 & 14.5265 \\
     &  & RMSE & \textbf{\textcolor{red}{17.2195}} & 18.6325 & 19.8707 & 18.8534 & 18.4689 & 20.3517 & 19.2234 & 20.5249 & 20.2049 & 18.7139 & \underline{\textcolor{blue}{17.5122}} & 19.4963 & 19.8336 & 18.9395 \\
     &  & MAPE & \textbf{\textcolor{red}{0.1747}} & 0.2507 & 0.2842 & 0.2492 & \underline{\textcolor{blue}{0.2148}} & 0.2546 & 0.2648 & 0.2831 & 0.2562 & 0.2582 & 0.2316 & 0.2837 & 0.2543 & 0.2353 \\
    \midrule
    \multirow{15}{*}{PEMSD4} & \multirow{3}{*}{96$\to$96} & MAE & \textbf{\textcolor{red}{26.4776}} & 29.6247 & 59.8549 & 64.2401 & \underline{\textcolor{blue}{26.6860}} & 49.0246 & 54.2296 & 74.9680 & 38.0736 & 41.5508 & 43.1233 & 44.3699 & 61.0265 & 32.6902 \\
     &  & RMSE & \textbf{\textcolor{red}{39.4029}} & 44.3683 & 82.6153 & 87.7110 & \underline{\textcolor{blue}{39.7370}} & 69.3914 & 74.4188 & 101.9582 & 53.4081 & 59.4928 & 59.5068 & 64.4355 & 80.7261 & 46.4117 \\
     &  & MAPE & \textbf{\textcolor{red}{0.2006}} & 0.2836 & 0.6127 & 0.7129 & \underline{\textcolor{blue}{0.2195}} & 0.5489 & 0.5666 & 0.8606 & 0.4147 & 0.4426 & 0.3569 & 0.3707 & 0.6928 & 0.2600 \\
    \cmidrule{2-17}
     & \multirow{3}{*}{96$\to$192} & MAE & \textbf{\textcolor{red}{28.1617}} & 33.8232 & 65.8329 & 67.7600 & \underline{\textcolor{blue}{28.2617}} & 53.1275 & 56.5315 & 81.9208 & 37.5710 & 43.7502 & 45.6979 & 46.7352 & 89.6164 & 36.7348 \\
     &  & RMSE & \textbf{\textcolor{red}{43.0837}} & 50.1289 & 90.8237 & 92.0194 & \underline{\textcolor{blue}{43.4020}} & 76.6645 & 77.8204 & 111.3066 & 53.7107 & 62.6010 & 67.6016 & 68.5116 & 112.3335 & 51.4468 \\
     &  & MAPE & \underline{\textcolor{blue}{0.2195}} & 0.3537 & 0.7415 & 0.8716 & \textbf{\textcolor{red}{0.2181}} & 0.6268 & 0.6523 & 1.0670 & 0.3973 & 0.4726 & 0.3845 & 0.4064 & 1.3163 & 0.3103 \\
    \cmidrule{2-17}
     & \multirow{3}{*}{96$\to$336} & MAE & \textbf{\textcolor{red}{29.3421}} & 32.3614 & 54.4556 & 56.2362 & \underline{\textcolor{blue}{30.2596}} & 48.4381 & 51.4121 & 64.0619 & 35.3102 & 40.8493 & 42.9502 & 42.4317 & 94.0688 & 39.4328 \\
     &  & RMSE & \textbf{\textcolor{red}{44.0975}} & 48.7913 & 77.6140 & 78.0632 & \underline{\textcolor{blue}{46.1562}} & 70.7553 & 71.7264 & 90.4817 & 52.0188 & 58.9058 & 60.5791 & 62.2943 & 118.7577 & 54.7326 \\
     &  & MAPE & \textbf{\textcolor{red}{0.2223}} & 0.3037 & 0.5675 & 0.6464 & \underline{\textcolor{blue}{0.2275}} & 0.5031 & 0.5342 & 0.7348 & 0.3261 & 0.4028 & 0.3528 & 0.3521 & 1.3383 & 0.3707 \\
    \cmidrule{2-17}
     & \multirow{3}{*}{96$\to$720} & MAE & \textbf{\textcolor{red}{31.5047}} & 38.3415 & 60.4930 & 62.8642 & \underline{\textcolor{blue}{31.7658}} & 54.2779 & 57.5695 & 74.5080 & 36.7164 & 44.4774 & 48.2680 & 47.5954 & 98.6730 & 41.7201 \\
     &  & RMSE & \textbf{\textcolor{red}{47.2747}} & 55.6071 & 85.7809 & 86.1355 & \underline{\textcolor{blue}{49.8363}} & 78.5152 & 79.8194 & 103.9511 & 54.1391 & 64.0994 & 67.4857 & 68.4360 & 123.6771 & 57.1715 \\
     &  & MAPE & \textbf{\textcolor{red}{0.2432}} & 0.3765 & 0.6471 & 0.7231 & \underline{\textcolor{blue}{0.2586}} & 0.5767 & 0.5978 & 0.8346 & 0.3395 & 0.4399 & 0.3907 & 0.3938 & 1.4187 & 0.4042 \\
    \cmidrule{2-17}
     & \multirow{3}{*}{Avg} & MAE & \textbf{\textcolor{red}{28.8715}} & 33.5377 & 60.1591 & 62.7751 & \underline{\textcolor{blue}{29.2433}} & 51.2170 & 54.9357 & 73.8647 & 36.9178 & 42.6569 & 45.0099 & 45.2831 & 85.8462 & 37.6445 \\
     &  & RMSE & \textbf{\textcolor{red}{43.4647}} & 49.7239 & 84.2085 & 85.9823 & \underline{\textcolor{blue}{44.7829}} & 73.8316 & 75.9463 & 101.9244 & 53.3192 & 61.2748 & 63.7933 & 65.9194 & 108.8736 & 52.4407 \\
     &  & MAPE & \textbf{\textcolor{red}{0.2214}} & 0.3294 & 0.6422 & 0.7385 & \underline{\textcolor{blue}{0.2309}} & 0.5639 & 0.5877 & 0.8743 & 0.3694 & 0.4395 & 0.3712 & 0.3808 & 1.1915 & 0.3363 \\
    \midrule
    \multirow{15}{*}{PEMSD8} & \multirow{3}{*}{96$\to$96} & MAE & \textbf{\textcolor{red}{23.7019}} & \underline{\textcolor{blue}{26.0281}} & 48.5769 & 47.2738 & 28.1961 & 38.7480 & 52.2431 & 64.8599 & 33.6599 & 35.6437 & 35.0671 & 44.3699 & 53.1667 & 29.7944 \\
     &  & RMSE & \textbf{\textcolor{red}{35.2860}} & \underline{\textcolor{blue}{39.8243}} & 67.8318 & 67.3346 & 40.6984 & 55.7459 & 72.1342 & 90.1120 & 47.4258 & 52.3985 & 48.9265 & 64.4355 & 69.7416 & 42.4747 \\
     &  & MAPE & \textbf{\textcolor{red}{0.1630}} & 0.2124 & 0.4085 & 0.3911 & 0.2072 & 0.3337 & 0.4109 & 0.5490 & 0.3116 & 0.3187 & 0.2645 & 0.3707 & 0.4914 & \underline{\textcolor{blue}{0.2046}} \\
    \cmidrule{2-17}
     & \multirow{3}{*}{96$\to$192} & MAE & \textbf{\textcolor{red}{25.6584}} & 28.9047 & 55.7625 & 50.1411 & \underline{\textcolor{blue}{28.4134}} & 44.2069 & 52.8373 & 66.6322 & 33.0867 & 39.9523 & 35.9352 & 46.7352 & 78.2649 & 33.6865 \\
     &  & RMSE & \textbf{\textcolor{red}{38.2914}} & 44.3403 & 78.0165 & 71.3803 & \underline{\textcolor{blue}{41.0260}} & 63.7083 & 72.9133 & 94.1912 & 47.0600 & 57.9489 & 50.1522 & 68.5116 & 97.6039 & 46.7601 \\
     &  & MAPE & \textbf{\textcolor{red}{0.1822}} & 0.2426 & 0.4670 & 0.4462 & \underline{\textcolor{blue}{0.2205}} & 0.4035 & 0.4377 & 0.5943 & 0.2739 & 0.3626 & 0.2986 & 0.4064 & 0.8610 & 0.2342 \\
    \cmidrule{2-17}
     & \multirow{3}{*}{96$\to$336} & MAE & \textbf{\textcolor{red}{25.4079}} & \underline{\textcolor{blue}{26.1305}} & 48.6164 & 44.2149 & 28.4132 & 40.4851 & 45.4043 & 59.0033 & 31.1443 & 33.2531 & 35.6601 & 42.4317 & 84.9112 & 34.2771 \\
     &  & RMSE & \textbf{\textcolor{red}{38.6768}} & \underline{\textcolor{blue}{40.7761}} & 68.7962 & 63.1292 & 41.7361 & 58.9178 & 63.1330 & 83.1697 & 45.2857 & 49.6175 & 49.7670 & 62.2943 & 105.0820 & 47.5110 \\
     &  & MAPE & \textbf{\textcolor{red}{0.1757}} & 0.2123 & 0.3902 & 0.3716 & \underline{\textcolor{blue}{0.2015}} & 0.3370 & 0.3555 & 0.5005 & 0.2418 & 0.2817 & 0.2894 & 0.3521 & 0.9324 & 0.2504 \\
    \cmidrule{2-17}
     & \multirow{3}{*}{96$\to$720} & MAE & \textbf{\textcolor{red}{29.0547}} & \underline{\textcolor{blue}{30.7741}} & 55.8387 & 51.6041 & 32.3941 & 45.2173 & 54.5998 & 65.1115 & 33.3680 & 40.2253 & 38.5227 & 47.5954 & 86.1753 & 36.6230 \\
     &  & RMSE & \textbf{\textcolor{red}{43.4398}} & 47.0546 & 77.6047 & 70.7531 & \underline{\textcolor{blue}{46.8004}} & 64.7816 & 74.7977 & 90.9112 & 49.0796 & 58.1527 & 53.3418 & 68.4360 & 106.6188 & 50.2912 \\
     &  & MAPE & \textbf{\textcolor{red}{0.2022}} & 0.2507 & 0.4527 & 0.4362 & \underline{\textcolor{blue}{0.2311}} & 0.3881 & 0.4385 & 0.5632 & 0.2514 & 0.3395 & 0.2996 & 0.3938 & 0.9458 & 0.2640 \\
    \cmidrule{2-17}
     & \multirow{3}{*}{Avg} & MAE & \textbf{\textcolor{red}{25.9557}} & \underline{\textcolor{blue}{27.9594}} & 52.1986 & 48.3085 & 29.3542 & 42.1643 & 51.2711 & 63.9017 & 32.8147 & 37.2686 & 36.2963 & 45.2831 & 75.6295 & 33.5953 \\
     &  & RMSE & \textbf{\textcolor{red}{38.9235}} & 42.9988 & 73.0623 & 68.1493 & \underline{\textcolor{blue}{42.5652}} & 60.7884 & 70.7446 & 89.5960 & 47.2128 & 54.5294 & 50.5469 & 65.9194 & 94.7616 & 46.7593 \\
     &  & MAPE & \textbf{\textcolor{red}{0.1808}} & 0.2295 & 0.4296 & 0.4113 & \underline{\textcolor{blue}{0.2151}} & 0.3656 & 0.4107 & 0.5518 & 0.2697 & 0.3256 & 0.2880 & 0.3808 & 0.8076 & 0.2383 \\
    \midrule
    % check从此开始
    \multirow{10}{*}{KnowAir} & \multirow{2}{*}{96$\to$96} & MAE & \textbf{\textcolor{red}{17.3852}} & 17.9307 & 18.2672 & 17.6891 & 22.4962 & \underline{\textcolor{blue}{17.5663}} & 17.9298 & 22.5112 & 20.0175 & 17.9527 & 17.6064 & 17.8256 & 18.2087 & 23.8860 \\
     &  & RMSE & \textbf{\textcolor{red}{24.6230}} & 24.6957 & 25.5774 & \underline{\textcolor{blue}{24.6585}} & 29.0041 & 25.2919 & 24.8181 & 41.6058 & 27.0068 & 24.9152 & 24.9475 & 25.0624 & 25.0230 & 30.5861 \\
    \cmidrule{2-17}
     & \multirow{2}{*}{96$\to$192} & MAE & \textbf{\textcolor{red}{17.3225}} & 17.9721 & 18.0976 & 17.6282 & 20.9691 & 17.5456 & 17.9437 & 18.7603 & 18.9868 & 17.8147 & \underline{\textcolor{blue}{17.4808}} & 17.8240 & 18.0729 & 21.4672 \\
     &  & RMSE & \textbf{\textcolor{red}{24.3828}} & 24.5736 & 25.3653 & 24.6629 & 27.3749 & 25.1842 & 24.5971 & 26.2448 & 26.3943 & 24.5735 & \underline{\textcolor{blue}{24.5267}} & 24.9554 & 25.4802 & 28.0253 \\
    \cmidrule{2-17}
     & \multirow{2}{*}{96$\to$336} & MAE & \textbf{\textcolor{red}{17.2260}} & 17.7218 & 17.7629 & 17.4451 & 19.4870 & 17.4568 & 17.5529 & 19.0424 & 20.9541 & 17.6995 & \underline{\textcolor{blue}{17.2968}} & 17.7196 & 17.7643 & 18.9041 \\
     &  & RMSE & \underline{\textcolor{blue}{24.0268}} & 24.2959 & 25.0730 & 24.3473 & 25.7854 & 25.1676 & 24.2337 & 27.1777 & 27.7828 & 24.3709 & \textbf{\textcolor{red}{24.0142}} & 24.8347 & 24.8325 & 25.3368 \\
    \cmidrule{2-17}
     & \multirow{2}{*}{96$\to$720} & MAE & \textbf{\textcolor{red}{16.4709}} & 17.7279 & 17.5585 & 17.0822 & 17.3846 & 17.1349 & 17.4344 & 18.0660 & 17.6544 & 17.6544 & \underline{\textcolor{blue}{16.5393}} & 17.4469 & 17.7880 & 17.1537 \\
     &  & RMSE & \underline{\textcolor{blue}{22.6527}} & 23.7959 & 24.2717 & 23.3701 & 23.3731 & 24.0558 & 23.5836 & 25.0577 & 23.8745 & 23.8745 & \textbf{\textcolor{red}{22.5549}} & 24.1019 & 23.7451 & 23.2010 \\
    \cmidrule{2-17}
     & \multirow{2}{*}{Avg} & MAE & \textbf{\textcolor{red}{17.1012}} & 17.8381 & 17.9216 & 17.4612 & 20.0842 & 17.4259 & 17.7152 & 19.5950 & 19.4032 & 17.7803 & \underline{\textcolor{blue}{17.2308}} & 17.7040 & 17.9585 & 20.3528 \\
     &  & RMSE & \textbf{\textcolor{red}{23.9213}} & 24.3403 & 25.0719 & 24.2597 & 26.3844 & 24.9249 & 24.3081 & 30.0215 & 26.2646 & 24.4335 & \underline{\textcolor{blue}{24.0108}} & 24.7386 & 24.7702 & 26.7873 \\
    \midrule
    \multirow{10}{*}{NREL} & \multirow{2}{*}{96$\to$96} & MAE & \textbf{\textcolor{red}{4.5565}} & 5.1207 & 5.4818 & 5.1535 & 5.4195 & \underline{\textcolor{blue}{5.0372}} & 5.3392 & 9.4901 & 5.6849 & 5.1281 & 5.3626 & 5.1176 & 9.1364 & 5.3306 \\
     &  & RMSE & \textbf{\textcolor{red}{6.2684}} & \underline{\textcolor{blue}{6.9011}} & 7.7656 & 6.9684 & 7.1565 & 7.2262 & 7.2574 & 13.1642 & 7.7404 & 6.9295 & 7.1660 & 7.2159 & 11.6403 & 7.0399 \\
    \cmidrule{2-17}
     & \multirow{2}{*}{96$\to$192} & MAE & \textbf{\textcolor{red}{5.2324}} & 5.7407 & 6.1818 & 5.7250 & 6.0108 & 5.7116 & 6.5975 & 6.5524 & 6.2416 & 5.8240 & \underline{\textcolor{blue}{5.6277}} & 5.7644 & 10.4102 & 6.8083 \\
     &  & RMSE & \textbf{\textcolor{red}{7.1398}} & 7.6061 & 8.4362 & 7.5996 & 7.8233 & 8.0476 & 8.7240 & 8.8330 & 8.5164 & 7.6355 & \underline{\textcolor{blue}{7.5718}} & 7.8733 & 13.3419 & 8.4299 \\
    \cmidrule{2-17}
     & \multirow{2}{*}{96$\to$336} & MAE & \textbf{\textcolor{red}{5.5272}} & 6.2035 & 6.5672 & \underline{\textcolor{blue}{6.1396}} & 6.4648 & 6.1598 & 6.3071 & 8.1089 & 6.9887 & 6.2007 & 6.2246 & 6.2760 & 10.0695 & 6.5919 \\
     &  & RMSE & \textbf{\textcolor{red}{7.5120}} & 8.0804 & 8.7676 & 8.0232 & 8.2357 & 8.4560 & 8.2819 & 10.8341 & 9.0983 & \underline{\textcolor{blue}{8.0130}} & 8.0936 & 8.4152 & 12.6833 & 8.2869 \\
    \cmidrule{2-17}
     & \multirow{2}{*}{96$\to$720} & MAE & \textbf{\textcolor{red}{5.9094}} & 6.7768 & 6.9916 & 6.7375 & 6.7876 & 6.7057 & 7.0396 & 7.9473 & 6.5956 & \underline{\textcolor{blue}{6.5771}} & 6.6764 & 6.8421 & 9.3011 & 7.2086 \\
     &  & RMSE & \textbf{\textcolor{red}{7.7697}} & 8.6619 & 9.2444 & 8.6542 & 8.5488 & 9.0519 & 9.0565 & 10.4567 & 8.6828 & \underline{\textcolor{blue}{8.3300}} & 8.5893 & 9.0278 & 12.2993 & 8.7554 \\
    \cmidrule{2-17}
     & \multirow{2}{*}{Avg} & MAE & \textbf{\textcolor{red}{5.3064}} & 5.9604 & 6.3056 & 5.9389 & 6.1707 & \underline{\textcolor{blue}{5.9036}} & 6.3209 & 8.0247 & 6.3777 & 5.9325 & 5.9728 & 6.0000 & 9.7293 & 6.4848 \\
     &  & RMSE & \textbf{\textcolor{red}{7.1725}} & 7.8124 & 8.5534 & 7.8114 & 7.9411 & 8.1954 & 8.3300 & 10.8220 & 8.5095 & \underline{\textcolor{blue}{7.7270}} & 7.8552 & 8.1331 & 12.4912 & 8.1280 \\
    \midrule
    \multirow{10}{*}{ETTh1} & \multirow{2}{*}{96$\to$96} & MAE & \underline{\textcolor{blue}{0.3906}} & 0.4154 & 0.4080 & \textbf{\textcolor{red}{0.3878}} & 0.5777 & 0.3941 & 0.4108 & 0.4515 & 0.8154 & 0.4026 & 0.4001 & 0.3945 & 0.6609 & 0.5585 \\
     &  & MSE & 0.3854 & 0.3991 & 0.3950 & \textbf{\textcolor{red}{0.3755}} & 0.6232 & 0.3804 & 0.3986 & 0.4583 & 0.9806 & 0.3843 & 0.3877 & \underline{\textcolor{blue}{0.3791}} & 0.8560 & 0.6221 \\
    \cmidrule{2-17}
     & \multirow{2}{*}{96$\to$192} & MAE & 0.4267 & 0.4514 & 0.4331 & \textbf{\textcolor{red}{0.4177}} & 0.6219 & \underline{\textcolor{blue}{0.4225}} & 0.4361 & 0.4345 & 0.7893 & 0.4302 & 0.4260 & 0.4236 & 0.6266 & 0.6635 \\
     &  & MSE & 0.4408 & 0.4624 & 0.4406 & \textbf{\textcolor{red}{0.4231}} & 0.6941 & 0.4362 & 0.4450 & 0.4418 & 1.0025 & 0.4363 & 0.4364 & \underline{\textcolor{blue}{0.4278}} & 0.7957 & 0.7638 \\
    \cmidrule{2-17}
     & \multirow{2}{*}{96$\to$336} & MAE & \textbf{\textcolor{red}{0.4375}} & 0.4565 & 0.4661 & \underline{\textcolor{blue}{0.4402}} & 0.6434 & 0.4443 & 0.4522 & 0.4892 & 0.7794 & 0.4546 & 0.4548 & 0.4476 & 0.6385 & 0.7591 \\
     &  & MSE & \textbf{\textcolor{red}{0.4551}} & 0.4898 & 0.4992 & \underline{\textcolor{blue}{0.4675}} & 0.7065 & 0.4785 & 0.4804 & 0.5374 & 0.9476 & 0.4801 & 0.4869 & 0.4738 & 0.7905 & 0.9097 \\
    \cmidrule{2-17}
     & \multirow{2}{*}{96$\to$720} & MAE & 0.4904 & 0.4677 & 0.4813 & \textbf{\textcolor{red}{0.4571}} & 0.7611 & \underline{\textcolor{blue}{0.4623}} & 0.4750 & 0.4711 & 0.7967 & 0.4761 & 0.5069 & 0.4764 & 0.6831 & 0.7770 \\
     &  & MSE & 0.5053 & \textbf{\textcolor{red}{0.4689}} & 0.4974 & \underline{\textcolor{blue}{0.4706}} & 0.8971 & 0.4759 & 0.4858 & 0.4819 & 0.9713 & 0.4799 & 0.5250 & 0.4868 & 0.8545 & 0.9516 \\
    \cmidrule{2-17}
     & \multirow{2}{*}{Avg} & MAE & 0.4363 & 0.4477 & 0.4471 & \textbf{\textcolor{red}{0.4257}} & 0.6510 & \underline{\textcolor{blue}{0.4308}} & 0.4435 & 0.4616 & 0.7952 & 0.4409 & 0.4470 & 0.4355 & 0.6523 & 0.6895 \\
     &  & MSE & 0.4467 & 0.4551 & 0.4581 & \textbf{\textcolor{red}{0.4342}} & 0.7302 & 0.4428 & 0.4525 & 0.4799 & 0.9755 & 0.4452 & 0.4590 & \underline{\textcolor{blue}{0.4419}} & 0.8242 & 0.8118 \\
    \bottomrule
    \end{tabular}%
  }
\end{table*}

\begin{table*}[htbp]
  \centering
  % \caption{Detailed experimental results on all datasets and horizons.}
  \label{tab:detailed_results_2}
  \resizebox{\textwidth}{!}{%
    \begin{tabular}{llccccccccccccccc}
    \toprule
    \textbf{Benchmark} & \textbf{In-Out} & \textbf{Metric} & \textbf{STM3} & \textbf{DST-Mamba} & \textbf{SST} & \textbf{MixMamba} & \textbf{STGMamba} & \textbf{DGraFormer} & \textbf{FilterTS} & \textbf{AMD} & \textbf{TimeMixer++} & \textbf{iTransformer} & \textbf{STGM} & \textbf{PatchTST} & \textbf{PromptST} & \textbf{Informer} \\
    \midrule
    \multirow{10}{*}{Electricity} & \multirow{2}{*}{96$\to$96} & MAE & \textbf{\textcolor{red}{0.2317}} & \underline{\textcolor{blue}{0.2433}} & 0.2608 & 0.2844 & 0.3782 & 0.2582 & 0.2622 & 0.2789 & 0.4618 & 0.2438 & 0.2705 & 0.2520 & 0.6729 & 0.3849 \\
     &  & MSE & \textbf{\textcolor{red}{0.1397}} & \underline{\textcolor{blue}{0.1454}} & 0.1770 & 0.1983 & 0.3017 & 0.1728 & 0.1604 & 0.1855 & 0.4102 & 0.1504 & 0.1925 & 0.1655 & 0.8033 & 0.2925 \\
    \cmidrule{2-17}
     & \multirow{2}{*}{96$\to$192} & MAE & \textbf{\textcolor{red}{0.2434}} & 0.2630 & 0.2666 & 0.2920 & 0.3999 & 0.2687 & 0.2715 & 0.2782 & 0.4865 & \underline{\textcolor{blue}{0.2544}} & 0.2775 & 0.2604 & 0.6751 & 0.3811 \\
     &  & MSE & \textbf{\textcolor{red}{0.1519}} & 0.1632 & 0.1828 & 0.2014 & 0.3194 & 0.1839 & 0.1745 & 0.1856 & 0.4361 & \underline{\textcolor{blue}{0.1626}} & 0.1976 & 0.1741 & 0.8003 & 0.2901 \\
    \cmidrule{2-17}
     & \multirow{2}{*}{96$\to$336} & MAE & \textbf{\textcolor{red}{0.2609}} & 0.2855 & 0.2833 & 0.3047 & 0.3916 & 0.2789 & 0.2878 & 0.3157 & 0.4417 & \underline{\textcolor{blue}{0.2728}} & 0.2901 & 0.2768 & 0.6771 & 0.3789 \\
     &  & MSE & \textbf{\textcolor{red}{0.1685}} & 0.1873 & 0.1992 & 0.2152 & 0.3185 & 0.1958 & 0.1919 & 0.2246 & 0.3824 & \underline{\textcolor{blue}{0.1793}} & 0.2074 & 0.1899 & 0.7987 & 0.2875 \\
    \cmidrule{2-17}
     & \multirow{2}{*}{96$\to$720} & MAE & \textbf{\textcolor{red}{0.3004}} & \underline{\textcolor{blue}{0.3030}} & 0.3165 & 0.3370 & 0.4144 & 0.3151 & 0.3254 & 0.3940 & 0.4512 & 0.3063 & 0.3219 & 0.3116 & 0.6790 & 0.3910 \\
     &  & MSE & \underline{\textcolor{blue}{0.2130}} & \textbf{\textcolor{red}{0.2071}} & 0.2392 & 0.2571 & 0.3452 & 0.2410 & 0.2371 & 0.3231 & 0.3906 & 0.2166 & 0.2421 & 0.2306 & 0.7891 & 0.3117 \\
    \cmidrule{2-17}
     & \multirow{2}{*}{Avg} & MAE & \textbf{\textcolor{red}{0.2591}} & 0.2737 & 0.2818 & 0.3045 & 0.3960 & 0.2802 & 0.2867 & 0.3167 & 0.4603 & \underline{\textcolor{blue}{0.2693}} & 0.2900 & 0.2752 & 0.6760 & 0.3840 \\
     &  & MSE & \textbf{\textcolor{red}{0.1683}} & \underline{\textcolor{blue}{0.1758}} & 0.1996 & 0.2180 & 0.3212 & 0.1984 & 0.1910 & 0.2297 & 0.4048 & 0.1772 & 0.2099 & 0.1900 & 0.7979 & 0.2955 \\  
    \midrule
    \multirow{10}{*}{Milan-SMS} & \multirow{2}{*}{12$\to$12} & MAE & \textbf{\textcolor{red}{13.5682}} & 20.7363 & 20.7165 & 17.3188 & \underline{\textcolor{blue}{15.4815}} & 19.5202 & 19.6901 & 18.5099 & 25.3473 & 15.7635 & 16.1515 & 21.1731 & 18.3259 & 22.1390 \\
     &  & RMSE & \textbf{\textcolor{red}{32.3681}} & 43.9357 & 45.4070 & 38.9703 & \underline{\textcolor{blue}{33.3397}} & 45.2224 & 43.3817 & 40.2651 & 47.7130 & 34.7501 & 34.4700 & 45.8881 & 42.2098 & 45.9527 \\
    \cmidrule{2-17}
     & \multirow{2}{*}{12$\to$24} & MAE & \textbf{\textcolor{red}{13.1261}} & 15.6559 & 16.6609 & 15.3345 & 18.9670 & 15.3067 & 16.8569 & 16.7511 & 30.3549 & \underline{\textcolor{blue}{14.1495}} & 15.1564 & 15.5538 & 21.6856 & 17.7760 \\
     &  & RMSE & \textbf{\textcolor{red}{32.4810}} & 35.4681 & 39.1353 & 35.4720 & 39.1891 & 37.0934 & 38.7699 & 37.9472 & 53.8839 & \underline{\textcolor{blue}{32.8416}} & 33.9167 & 37.1085 & 46.3612 & 43.3241 \\
    \cmidrule{2-17}
     & \multirow{2}{*}{12$\to$48} & MAE & \textbf{\textcolor{red}{14.8748}} & 16.3250 & 18.0522 & 16.2492 & 20.0540 & 16.8893 & 17.4744 & 17.3886 & 32.2965 & \underline{\textcolor{blue}{15.4441}} & 16.6851 & 16.5870 & 21.8229 & 20.8325 \\
     &  & RMSE & 35.6124 & 36.2147 & 41.5880 & 36.9891 & 37.2296 & 40.2865 & 39.8521 & 38.7442 & 57.8410 & \underline{\textcolor{blue}{35.3661}} & \textbf{\textcolor{red}{35.0240}} & 38.7628 & 46.3908 & 47.5510 \\
    \cmidrule{2-17}
     & \multirow{2}{*}{12$\to$96} & MAE & \textbf{\textcolor{red}{17.7674}} & 19.0976 & 19.4453 & \underline{\textcolor{blue}{17.7906}} & 23.3057 & 18.4115 & 18.5115 & 18.6220 & 31.8196 & 17.8514 & 20.4586 & 18.0943 & 23.1533 & 19.6335 \\
     &  & RMSE & \underline{\textcolor{blue}{39.5692}} & 41.5741 & 42.8064 & \textbf{\textcolor{red}{39.4189}} & 52.8707 & 41.8188 & 41.6089 & 41.3531 & 59.4501 & 40.2164 & 41.8184 & 40.5338 & 48.6886 & 46.2258 \\
    \cmidrule{2-17}
     & \multirow{2}{*}{Avg} & MAE & \textbf{\textcolor{red}{14.8341}} & 17.9537 & 18.7187 & 16.6733 & 19.4520 & 17.5319 & 18.1332 & 17.8179 & 29.9546 & \underline{\textcolor{blue}{15.8021}} & 17.1129 & 17.8520 & 21.2469 & 20.0953 \\
     &  & RMSE & \textbf{\textcolor{red}{35.0077}} & 39.2982 & 42.2342 & 37.7126 & 40.6573 & 41.1053 & 40.9032 & 39.5774 & 54.7220 & \underline{\textcolor{blue}{35.7936}} & 36.3073 & 40.5733 & 45.9126 & 45.7634 \\
    \midrule
    \multirow{10}{*}{Milan-Call} & \multirow{2}{*}{12$\to$12} & MAE & \textbf{\textcolor{red}{9.3960}} & 16.2223 & 14.5397 & 11.5841 & \underline{\textcolor{blue}{10.5271}} & 12.2304 & 14.8322 & 14.9238 & 22.3670 & 10.9994 & 12.3763 & 13.0503 & 16.2266 & 23.4324 \\
     &  & RMSE & \textbf{\textcolor{red}{24.2080}} & 37.3671 & 34.9916 & 29.7563 & \underline{\textcolor{blue}{25.2022}} & 31.2138 & 35.1099 & 35.6374 & 42.6412 & 27.1594 & 29.7759 & 32.5382 & 38.8956 & 46.0955 \\
    \cmidrule{2-17}
     & \multirow{2}{*}{12$\to$24} & MAE & \textbf{\textcolor{red}{9.6919}} & 11.4065 & 13.4185 & 11.1157 & 18.0898 & 11.6663 & 12.5772 & 12.7916 & 30.4808 & \underline{\textcolor{blue}{10.8446}} & 11.5510 & 11.1859 & 21.5256 & 20.4622 \\
     &  & RMSE & \textbf{\textcolor{red}{26.7106}} & 28.4912 & 34.0505 & 29.7261 & 37.2061 & 32.1039 & 32.9354 & 33.0256 & 51.8304 & \underline{\textcolor{blue}{28.2425}} & 29.1496 & 30.8958 & 47.3518 & 45.7000 \\
    \cmidrule{2-17}
     & \multirow{2}{*}{12$\to$48} & MAE & \textbf{\textcolor{red}{12.7806}} & 13.9591 & 15.4446 & \underline{\textcolor{blue}{13.0915}} & 19.1010 & 13.7448 & 14.7169 & 15.0045 & 32.6349 & 13.1027 & 15.6470 & 13.1990 & 22.0167 & 22.1299 \\
     &  & RMSE & \underline{\textcolor{blue}{33.3597}} & 34.3617 & 38.8031 & 33.9428 & 37.8647 & 37.2800 & 37.4317 & 38.1138 & 57.7679 & \textbf{\textcolor{red}{32.6753}} & 35.8008 & 35.5954 & 47.2407 & 47.1526 \\
    \cmidrule{2-17}
     & \multirow{2}{*}{12$\to$96} & MAE & 13.7592 & 13.9522 & 14.7760 & 14.4685 & 22.2802 & \textbf{\textcolor{red}{12.8756}} & 14.6219 & 14.6077 & 32.2367 & 14.3141 & 15.6470 & \underline{\textcolor{blue}{13.0311}} & 22.1631 & 18.4083 \\
     &  & RMSE & 34.9308 & \textbf{\textcolor{red}{34.2177}} & 37.8280 & \underline{\textcolor{blue}{34.5747}} & 52.0541 & 35.7589 & 37.2443 & 36.8914 & 58.4523 & 35.2890 & 35.8008 & 34.9155 & 47.6167 & 46.5335 \\
    \cmidrule{2-17}
     & \multirow{2}{*}{Avg} & MAE & \textbf{\textcolor{red}{11.4069}} & 13.8850 & 14.5447 & 12.5650 & 17.4995 & 12.6293 & 14.1871 & 14.3319 & 29.4298 & \underline{\textcolor{blue}{12.3152}} & 13.8053 & 12.6166 & 20.4830 & 21.1082 \\
     &  & RMSE & \textbf{\textcolor{red}{29.8023}} & 33.6094 & 36.4183 & 32.0000 & 38.0818 & 34.0892 & 35.6803 & 35.9171 & 52.6730 & \underline{\textcolor{blue}{30.8416}} & 32.6318 & 33.4862 & 45.2762 & 46.3704 \\
    \midrule
    \multirow{10}{*}{Milan-Internet} & \multirow{2}{*}{12$\to$12} & MAE & \textbf{\textcolor{red}{51.3435}} & 59.5043 & 64.9344 & 61.3548 & 103.6307 & \underline{\textcolor{blue}{56.9848}} & 64.2041 & 67.7419 & 173.6062 & 61.2549 & 73.1997 & 63.3876 & 87.7402 & 114.8223 \\
     &  & RMSE & \textbf{\textcolor{red}{127.3646}} & 145.5055 & 158.9911 & 156.9196 & 213.5057 & \underline{\textcolor{blue}{142.0015}} & 156.1836 & 158.4314 & 320.2741 & 148.7978 & 181.3639 & 155.2593 & 214.2488 & 273.9963 \\
    \cmidrule{2-17}
     & \multirow{2}{*}{12$\to$24} & MAE & 53.8845 & \textbf{\textcolor{red}{53.5506}} & 56.1467 & \underline{\textcolor{blue}{53.8419}} & 136.5017 & 55.0248 & 55.1874 & 61.2420 & 204.1067 & 54.5533 & 67.1227 & 54.8507 & 89.0389 & 109.4824 \\
     &  & RMSE & \underline{\textcolor{blue}{139.1029}} & \textbf{\textcolor{red}{137.6399}} & 145.5679 & 143.3453 & 274.6622 & 141.6689 & 145.0951 & 153.3095 & 369.7765 & 140.0878 & 152.6001 & 145.0302 & 227.0835 & 279.4567 \\
    \cmidrule{2-17}
     & \multirow{2}{*}{12$\to$48} & MAE & \textbf{\textcolor{red}{55.3451}} & \underline{\textcolor{blue}{56.4295}} & 62.8608 & 58.5078 & 132.3025 & 57.0619 & 59.5375 & 65.2182 & 226.5276 & 61.9121 & 78.1387 & 59.3071 & 103.8322 & 117.0214 \\
     &  & RMSE & \textbf{\textcolor{red}{143.7303}} & \underline{\textcolor{blue}{156.4483}} & 164.3439 & 157.4887 & 260.4425 & 156.7847 & 158.8084 & 166.3998 & 388.3850 & 167.3350 & 177.1002 & 159.3896 & 261.6330 & 279.8950 \\
    \cmidrule{2-17}
     & \multirow{2}{*}{12$\to$96} & MAE & 66.8640 & 63.9292 & 66.4582 & 62.2985 & 177.7212 & \underline{\textcolor{blue}{61.8319}} & 62.9079 & 78.5766 & 222.7269 & \textbf{\textcolor{red}{59.7755}} & 79.4025 & 63.6345 & 103.3742 & 132.0666 \\
     &  & RMSE & 168.5287 & \textbf{\textcolor{red}{161.6121}} & 174.7912 & 162.2111 & 353.7709 & 166.7952 & 164.1915 & 182.3926 & 383.7936 & \underline{\textcolor{blue}{161.7914}} & 183.2717 & 167.8802 & 234.2134 & 301.6224 \\
    \cmidrule{2-17}
     & \multirow{2}{*}{Avg} & MAE & \textbf{\textcolor{red}{56.8593}} & 58.3534 & 62.6000 & 59.0007 & 137.5390 & \underline{\textcolor{blue}{57.7259}} & 60.4592 & 68.1947 & 206.7418 & 59.3740 & 74.4659 & 60.2950 & 95.9964 & 118.3482 \\
     &  & RMSE & \textbf{\textcolor{red}{144.6816}} & \underline{\textcolor{blue}{150.3015}} & 160.9235 & 154.9912 & 275.5953 & 151.8126 & 156.0697 & 165.1333 & 365.5573 & 154.5030 & 173.5840 & 156.8898 & 234.2947 & 283.7426 \\
    \midrule
    \multicolumn{3}{l}{\textbf{1st Count}} & \textbf{\textcolor{red}{94}} & 6 & 0 & \underline{\textcolor{blue}{8}} & 1 & 1 & 0 & 0 & 0 & 2 & 3 & 0 & 0 & 0 \\
    \bottomrule
    \end{tabular}%
  }
\end{table*}

\subsection{Computation Efficiency Analysis}\label{appendix:computation_cost}

\begin{table*}[htbp]
  \centering
  \caption{Computation cost and performance comparison on the PEMSD8 dataset. OOM: Out-of-Memory of a single 24GB GPU.}
  \label{tab:performance}
  \resizebox{0.7\textwidth}{!}{%
    \begin{tabular}{lccccccc}
    \toprule
    \multirow{2}{*}{Model} & \multicolumn{2}{c}{Training} & \multicolumn{2}{c}{Inference} & \multicolumn{3}{c}{Performance Metric} \\
    \cmidrule(lr){2-3} \cmidrule(lr){4-5} \cmidrule(lr){6-8}
          & Time (s/ep) & Mem (GB) & Time (s/ep) & Mem (GB) & MAE   & RMSE  & MAPE \\
    \midrule
    Informer & 25.31 & 6.15  & 3.67  & 3.05  & 36.6230 & 50.2912 & 0.2640 \\
    PromptST& 35.82 & 4.47  & 4.23  & 1.70  & 86.1753 & 106.6188 & 0.9458 \\
    PatchTST & 22.33 & 3.69  & 2.67  & 0.96  & 47.5954 & 68.4360 & 0.3938 \\
    STGM & -   & OOM     & 17.99 & 10.55  & 38.5227 & 53.3418 & 0.2996 \\
    iTransformer & 9.65  & 1.79  & 1.90  & 0.41  & 40.2253 & 58.1527 & 0.3395 \\
    Timemixer++ & 25.60 & 0.36  & 3.83  & 0.20  & 33.3680 & 49.0796 & 0.2514 \\
    AMD & 233.28 & 2.23  & 26.46 & 0.25  & 65.1115 & 90.9112 & 0.5632 \\
    FilterTS & 6.52  & 1.27  & 1.77  & 0.56  & 54.5998 & 74.7977 & 0.4385 \\
    DGraFormer & 9.72  & 2.08  & 1.86  & 0.95  & 45.2173 & 64.7816 & 0.3881 \\
    STGMamba & 151.27 & 2.51  & 3.75  & 0.47  & 32.3941 & 46.8004 & 0.2311 \\
    MixMamba & -   & OOM     & 68.77 & 8.87  & 51.6041 & 70.7531 & 0.4362 \\
    SST & 8.70  & 0.73  & 1.88  & 0.22  & 55.8387 & 77.6047 & 0.4527 \\
    DST-Mamba & 15.13 & 2.64  & 2.26  & 1.35  & 30.7741 & 47.0546 & 0.2507 \\
    \textbf{\textit{STM3}}  & 142.56 & 15.22 & 7.23  & 1.39  & \textbf{29.0547} & \textbf{43.4398} & \textbf{0.2022} \\
    \bottomrule
    \end{tabular}%
  }
\end{table*}

\begin{table*}[t]
\centering
\caption{Efficiency Analysis on NREL dataset. \textbf{Panel A} compares overall training/inference costs. \textbf{Panel B} and \textbf{C} analyze scalability w.r.t Scale Number and Node Percentage during inference. \textbf{OOM}: Out-of-Memory of a single 24GB GPU.}
\label{tab:efficiency_combined}

% ==========================================
% Panel A: Top (Full Width)
% ==========================================
% \resizebox{1.0\linewidth}{!}{
\resizebox{0.5\linewidth}{!}{
\begin{tabular}{l cccc}
\toprule
\multicolumn{5}{c}{\textbf{Panel A: Overall Cost}} \\
\cmidrule(lr){1-5}
\multirow{2.5}{*}{\textbf{Method}} & \multicolumn{2}{c}{\textbf{Training}} & \multicolumn{2}{c}{\textbf{Inference}} \\
\cmidrule(lr){2-3} \cmidrule(lr){4-5}
 & Time (s) & Mem (GB) & Time (s) & Mem (GB) \\
\midrule
\textbf{STM3 (Ours)} & \textbf{56.27} & \textbf{12.18} & \textbf{3.05} & \textbf{0.56} \\
Vanilla Mamba & 122.70 & 18.19 & 10.21 & 1.23 \\
Vanilla Transformer & -- & OOM & 16.04 & 1.37 \\
\bottomrule
\end{tabular}
}

\vspace{2pt} % 微调上下间距

% ==========================================
% Panel B & C: Bottom (Side by Side, Fixed Height)
% ==========================================
\begin{minipage}{1.0\linewidth}
    % 设置统一的字体大小和小间距，确保不使用 resizebox 也能放得下
    \scriptsize 
    \fontsize{8pt}{10pt}\selectfont % 字号，行距
    \setlength{\tabcolsep}{2pt} % 稍微减小列间距以防溢出

    % --- Panel B: Left ---
    \begin{minipage}[t]{0.49\linewidth}
        \centering
        % 使用 tabular* 强制填满 minipage 的宽度，而不是缩放
        \begin{tabular*}{\linewidth}{@{\extracolsep{\fill}}l c cc}
        \toprule
        \multicolumn{4}{c}{\textbf{Panel B: Scalability (Scale Num)}} \\
        \cmidrule(lr){1-4}
        \textbf{Method} & \textbf{Scale} & \textbf{Time(s)} & \textbf{Mem(GB)} \\
        \midrule
        \multirow{3}{*}{\shortstack[l]{\textbf{STM3}\\\textbf{(Ours)}}} 
          & 2 & \textbf{2.44} & \textbf{0.34} \\
          & 4 & \textbf{3.05} & \textbf{0.56} \\
          & 6 & \textbf{3.98} & \textbf{0.79} \\
        \cmidrule{1-4}
        \multirow{3}{*}{\shortstack[l]{Vanilla\\Mamba}} 
          & 2 & 4.28 & 0.63 \\
          & 4 & 10.21 & 1.23 \\
          & 6 & 20.08 & 1.94 \\
        \cmidrule{1-4}
        \multirow{3}{*}{\shortstack[l]{Vanilla\\Trans.}} 
          & 2 & 7.83 & 1.04 \\
          & 4 & 16.04 & 1.37 \\
          & 6 & 26.80 & 1.80 \\
        \bottomrule
        \end{tabular*}
    \end{minipage}
    \hfill
    % --- Panel C: Right ---
    \begin{minipage}[t]{0.49\linewidth}
        \centering
        \begin{tabular*}{\linewidth}{@{\extracolsep{\fill}}l c cc}
        \toprule
        \multicolumn{4}{c}{\textbf{Panel C: Scalability (Node \%)}} \\
        \cmidrule(lr){1-4}
        \textbf{Method} & \textbf{Nodes} & \textbf{Time(s)} & \textbf{Mem(GB)} \\
        \midrule
        \multirow{3}{*}{\shortstack[l]{\textbf{STM3}\\\textbf{(Ours)}}} 
          & 33\% & \textbf{2.01} & \textbf{0.19} \\
          & 67\% & \textbf{2.70} & \textbf{0.38} \\
          & 100\% & \textbf{3.05} & \textbf{0.56} \\
        \cmidrule{1-4}
        \multirow{3}{*}{\shortstack[l]{Vanilla\\Mamba}} 
          & 33\% & 4.91 & 0.41 \\
          & 67\% & 7.37 & 0.82 \\
          & 100\% & 10.21 & 1.23 \\
        \cmidrule{1-4}
        \multirow{3}{*}{\shortstack[l]{Vanilla\\Trans.}} 
          & 33\% & 6.42 & 0.46 \\
          & 67\% & 11.28 & 0.91 \\
          & 100\% & 16.04 & 1.37 \\
        \bottomrule
        \end{tabular*}
    \end{minipage}
\end{minipage}

\end{table*}

\paragraph{Main Results and Comparative Analysis}
We conducted a comprehensive performance evaluation on the PEMSD8 dataset to assess both the predictive capability and the computational overhead of the proposed STM3. The experimental setup utilized a batch size of 64 with an input-output horizon of 96-720, implemented on a single NVIDIA RTX 3090 GPU. The comparative results against state-of-the-art baselines are summarized in Table~\ref{tab:performance}.
\textbf{As observed, STM3 achieves superior performance across all error metrics, recording the lowest MAE, RMSE, and MAPE.} Regarding computational cost, although the training time is higher than lightweight models like iTransformer, \textbf{this is a worthwhile investment derived from our intricate modeling of fine-grained multi-scale information and the co-training of multiple experts, which are essential for achieving such superior accuracy}. Furthermore, our top-1 MoE architecture successfully mitigates the computational overhead during deployment. By activating only a subset of experts, STM3 maintains a highly efficient inference speed (7.23 s/epoch), which is substantially faster than other complex baselines such as MixMamba (68.77 s), AMD (26.46 s), and STGM (17.99 s). \textbf{This indicates that STM3 strikes an optimal balance: it leverages complex multi-expert learning for high accuracy while ensuring the rapid inference capabilities required for real-time deployment through sparse conditional computation.}

\paragraph{Ablation on Computational Efficiency}
To further validate the efficiency of our \textbf{cross-scale shared Mamba design, Multiscale Mamba}, we compared STM3 against unoptimized variants where the shared backbone is replaced with scale-specific, non-shared modules. We constructed two baselines, \textit{Vanilla Mamba} and \textit{Vanilla Transformer}, and conducted experiments on the NREL dataset using a single 24GB GPU, as detailed in Table~\ref{tab:efficiency_combined}.
The results highlight the critical role of parameter sharing in preventing computational explosion. As shown in Panel A, \textbf{without the shared design, the computational burden increases drastically}: STM3 requires less than half the training time of Vanilla Mamba, whereas the Vanilla Transformer fails entirely due to Out-of-Memory (OOM) errors. In the inference phase, STM3 is $3.3\times$ and $5.2\times$ faster than the Mamba and Transformer variants, respectively. Panels B and C further illustrate the scalability of our method: \textbf{as the scale number and node percentage increase, the unoptimized baselines suffer from severe latency degradation}. Notably, at the largest scale setting, STM3 outperforms Vanilla Mamba by $5.0\times$ and Vanilla Transformer by $6.7\times$. This analysis confirms that without the shared-backbone strategy, the computational cost for modeling multiple scales would be prohibitive, thereby justifying \textbf{the design as essential for achieving the high performance-to-cost ratio}.

\subsection{Full Ablation Study}\label{appendix:full_ablation}

We provided the detailed ablation study results in Table~\ref{tab:full_ablation}.

\newpage

\begin{table*}[t]
\centering
\caption{Detailed ablation study results on all benchmarks and in-out settings.}
\label{tab:full_ablation}
\resizebox{\textwidth}{!}{
\setlength{\tabcolsep}{8pt} % Column spacing
\fontsize{8pt}{9pt}\selectfont % Font size, row spacing
\begin{tabular}{llc|ccccccc}
\toprule
Dataset & In-Out & Metric & \textbf{STM3} & \textbf{w/o DMoE} & \textbf{w/o CL} & \textbf{w/o Multiscale} & \textbf{w/o CA} & \textbf{vanilla Mamba} & \textbf{vanilla routing} \\
\midrule
\multirow{15}{*}{\textbf{METR\_LA}} & \multirow{3}{*}{96$\to$96} & MAE & 9.5837 & 9.8382 & 9.6152 & 10.0008 & 9.9661 & 9.7494 & 9.8707 \\
 &  & RMSE & 15.5923 & 16.1127 & 15.9298 & 16.1637 & 16.0445 & 15.7381 & 15.8448 \\
 &  & MAPE & 0.1772 & 0.1994 & 0.1852 & 0.1994 & 0.1863 & 0.1788 & 0.1844 \\
\cmidrule{2-10}
 & \multirow{3}{*}{96$\to$192} & MAE & 10.8418 & 11.1034 & 10.9452 & 11.2671 & 10.9578 & 10.9904 & 11.0194 \\
 &  & RMSE & 17.1490 & 17.3986 & 17.1629 & 17.4737 & 17.2431 & 17.1926 & 17.1250 \\
 &  & MAPE & 0.1770 & 0.1856 & 0.1819 & 0.1929 & 0.1839 & 0.1797 & 0.1823 \\
\cmidrule{2-10}
 & \multirow{3}{*}{96$\to$336} & MAE & 11.2723 & 11.4042 & 11.3807 & 11.6949 & 11.4242 & 11.3654 & 11.4956 \\
 &  & RMSE & 17.4188 & 17.7463 & 17.6777 & 18.0139 & 17.7412 & 17.5450 & 17.7698 \\
 &  & MAPE & 0.1662 & 0.1733 & 0.1708 & 0.1848 & 0.1744 & 0.1708 & 0.1777 \\
\cmidrule{2-10}
 & \multirow{3}{*}{96$\to$720} & MAE & 12.2057 & 12.3315 & 12.2151 & 12.5781 & 12.4860 & 12.3044 & 12.3620 \\
 &  & RMSE & 18.7180 & 19.0761 & 18.8201 & 19.4226 & 19.2825 & 18.8848 & 18.9400 \\
 &  & MAPE & 0.1783 & 0.1829 & 0.1796 & 0.1928 & 0.1914 & 0.1800 & 0.1833 \\
\cmidrule{2-10}
 & \multirow{3}{*}{\textbf{Avg}} & MAE & \textbf{10.9759} & 11.1693 & 11.0391 & 11.3852 & 11.2085 & 11.1024 & 11.1869 \\
 &  & RMSE & \textbf{17.2195} & 17.5834 & 17.3976 & 17.7685 & 17.5778 & 17.3401 & 17.4199 \\
 &  & MAPE & \textbf{0.1747} & 0.1853 & 0.1794 & 0.1925 & 0.1840 & 0.1773 & 0.1819 \\
\hline
\multirow{15}{*}{\textbf{PEMSD4}} & \multirow{3}{*}{96$\to$96} & MAE & 26.4776 & 26.7826 & 26.8520 & 29.3333 & 26.6780 & 26.7072 & 26.4794 \\
 &  & RMSE & 39.4029 & 39.3982 & 39.7115 & 43.1994 & 39.5218 & 39.5162 & 39.6909 \\
 &  & MAPE & 0.2006 & 0.2109 & 0.2016 & 0.2187 & 0.2178 & 0.2159 & 0.2020 \\
\cmidrule{2-10}
 & \multirow{3}{*}{96$\to$192} & MAE & 28.1617 & 29.8841 & 29.3542 & 32.9290 & 28.2848 & 29.9873 & 28.7089 \\
 &  & RMSE & 43.0837 & 43.6557 & 43.2633 & 47.5703 & 43.5161 & 43.9315 & 44.3701 \\
 &  & MAPE & 0.2195 & 0.2500 & 0.2457 & 0.3042 & 0.2369 & 0.2440 & 0.2344 \\
\cmidrule{2-10}
 & \multirow{3}{*}{96$\to$336} & MAE & 29.3421 & 31.8155 & 30.3526 & 34.9542 & 30.9095 & 29.7719 & 29.8570 \\
 &  & RMSE & 44.0975 & 46.4858 & 45.0092 & 51.1549 & 45.3155 & 44.2198 & 44.5379 \\
 &  & MAPE & 0.2223 & 0.2619 & 0.2330 & 0.2729 & 0.2629 & 0.2240 & 0.2466 \\
\cmidrule{2-10}
 & \multirow{3}{*}{96$\to$720} & MAE & 31.5047 & 33.9329 & 33.2488 & 38.9721 & 32.8790 & 33.8057 & 32.0034 \\
 &  & RMSE & 47.2747 & 49.6139 & 48.9331 & 56.5682 & 48.2145 & 49.7633 & 47.2880 \\
 &  & MAPE & 0.2432 & 0.2683 & 0.2662 & 0.3231 & 0.2672 & 0.2667 & 0.2437 \\
\cmidrule{2-10}
 & \multirow{3}{*}{\textbf{Avg}} & MAE & \textbf{28.8715} & 30.6038 & 29.9519 & 34.0472 & 29.6878 & 30.0680 & 29.2622 \\
 &  & RMSE & \textbf{43.4647} & 44.7884 & 44.2293 & 49.6232 & 44.1420 & 44.3577 & 43.9717 \\
 &  & MAPE & \textbf{0.2214} & 0.2478 & 0.2366 & 0.2797 & 0.2462 & 0.2377 & 0.2317 \\
\hline
\multirow{15}{*}{\textbf{PEMSD8}} & \multirow{3}{*}{96$\to$96} & MAE & 23.7019 & 26.1699 & 24.1525 & 27.0864 & 24.1031 & 25.1331 & 23.8031 \\
 &  & RMSE & 35.2860 & 37.8222 & 35.3826 & 38.8021 & 35.4405 & 36.5617 & 35.6289 \\
 &  & MAPE & 0.1630 & 0.1848 & 0.1748 & 0.1797 & 0.1648 & 0.1750 & 0.1686 \\
\cmidrule{2-10}
 & \multirow{3}{*}{96$\to$192} & MAE & 25.6584 & 30.2553 & 26.9548 & 31.4160 & 26.5549 & 27.5244 & 26.3056 \\
 &  & RMSE & 38.2914 & 42.8754 & 39.0545 & 44.9798 & 38.6946 & 39.4313 & 39.7671 \\
 &  & MAPE & 0.1822 & 0.2291 & 0.1958 & 0.2156 & 0.1874 & 0.2059 & 0.1879 \\
\cmidrule{2-10}
 & \multirow{3}{*}{96$\to$336} & MAE & 25.4079 & 27.4512 & 26.6475 & 31.3755 & 25.5431 & 30.1635 & 25.4801 \\
 &  & RMSE & 38.6768 & 39.7453 & 39.1149 & 44.6829 & 38.9659 & 42.9063 & 37.4170 \\
 &  & MAPE & 0.1757 & 0.1982 & 0.1903 & 0.2215 & 0.1814 & 0.2202 & 0.1842 \\
\cmidrule{2-10}
 & \multirow{3}{*}{96$\to$720} & MAE & 29.0547 & 31.7463 & 31.5428 & 34.8324 & 31.9574 & 31.9794 & 33.9061 \\
 &  & RMSE & 43.4398 & 45.4341 & 45.4574 & 49.1840 & 45.5711 & 45.4366 & 48.1937 \\
 &  & MAPE & 0.2022 & 0.2206 & 0.2193 & 0.2415 & 0.2266 & 0.2399 & 0.2474 \\
\cmidrule{2-10}
 & \multirow{3}{*}{\textbf{Avg}} & MAE & \textbf{25.9557} & 28.9057 & 27.3244 & 31.1776 & 27.0396 & 28.7001 & 27.3737 \\
 &  & RMSE & \textbf{38.9235} & 41.4693 & 39.7524 & 44.4122 & 39.6680 & 41.0840 & 40.2517 \\
 &  & MAPE & \textbf{0.1808} & 0.2082 & 0.1951 & 0.2146 & 0.1901 & 0.2103 & 0.1970 \\
\hline
\multirow{10}{*}{\textbf{KnowAir}} & \multirow{2}{*}{96$\to$96} & MAE & 17.3852 & 17.4208 & 17.3929 & 17.4382 & 17.4136 & 17.4345 & 17.3920 \\
 &  & RMSE & 24.6230 & 24.6471 & 24.6446 & 24.6829 & 24.6784 & 24.7835 & 24.8873 \\
\cmidrule{2-10}
 & \multirow{2}{*}{96$\to$192} & MAE & 17.3225 & 17.3438 & 17.3563 & 17.3277 & 17.3775 & 17.5149 & 17.3238 \\
 &  & RMSE & 24.3828 & 24.3912 & 24.6284 & 24.4752 & 24.3828 & 24.4199 & 24.3837 \\
\cmidrule{2-10}
 & \multirow{2}{*}{96$\to$336} & MAE & 17.2260 & 17.2616 & 17.2532 & 17.2776 & 17.4861 & 17.3232 & 17.2415 \\
 &  & RMSE & 24.0268 & 24.4627 & 24.2361 & 24.3016 & 24.1854 & 24.3305 & 24.1646 \\
\cmidrule{2-10}
 & \multirow{2}{*}{96$\to$720} & MAE & 16.4709 & 17.0733 & 16.8359 & 16.5822 & 16.6858 & 16.6456 & 17.2454 \\
 &  & RMSE & 22.6527 & 23.0166 & 22.9358 & 23.2466 & 22.8515 & 22.7730 & 23.0495 \\
\cmidrule{2-10}
 & \multirow{2}{*}{\textbf{Avg}} & MAE & \textbf{17.1012} & 17.2749 & 17.2096 & 17.1564 & 17.2407 & 17.2296 & 17.3007 \\
 &  & RMSE & \textbf{23.9213} & 24.1294 & 24.1112 & 24.1766 & 24.0245 & 24.0767 & 24.1213 \\
\bottomrule
\end{tabular}
}
\end{table*}

\begin{table*}[t]
\centering
% \caption{Detailed Ablation Study Results (Part II). Only the best results in the \textbf{Avg} rows are highlighted in \textbf{bold}.}
\label{tab:ablation_part2}
\resizebox{\textwidth}{!}{
\setlength{\tabcolsep}{8pt} % Column spacing
\fontsize{8pt}{9pt}\selectfont % Font size, row spacing
\begin{tabular}{llc|ccccccc}
\toprule
Dataset & In-Out & Metric & \textbf{STM3} & \textbf{w/o DMoE} & \textbf{w/o CL} & \textbf{w/o Multiscale} & \textbf{w/o CA} & \textbf{vanilla Mamba} & \textbf{vanilla routing} \\
\midrule
\multirow{10}{*}{\textbf{NREL}} & \multirow{2}{*}{96$\to$96} & MAE & 4.5565 & 4.6801 & 4.8041 & 4.6973 & 4.6539 & 4.6764 & 4.5770 \\
 &  & RMSE & 6.2684 & 6.3461 & 6.6098 & 6.5367 & 6.4465 & 6.4330 & 6.3642 \\
\cmidrule{2-10}
 & \multirow{2}{*}{96$\to$192} & MAE & 5.2324 & 5.2687 & 5.2908 & 5.4174 & 5.4019 & 5.2395 & 5.2880 \\
 &  & RMSE & 7.1398 & 7.2723 & 7.1284 & 7.4555 & 7.2987 & 7.1896 & 7.1910 \\
\cmidrule{2-10}
 & \multirow{2}{*}{96$\to$336} & MAE & 5.5272 & 5.5294 & 5.8230 & 5.7449 & 5.8836 & 5.6130 & 5.6628 \\
 &  & RMSE & 7.5120 & 7.5194 & 7.7534 & 7.8484 & 7.7861 & 7.5305 & 7.5182 \\
\cmidrule{2-10}
 & \multirow{2}{*}{96$\to$720} & MAE & 5.9094 & 5.9533 & 6.0978 & 5.9682 & 6.3637 & 6.2008 & 6.2668 \\
 &  & RMSE & 7.7697 & 7.7891 & 7.8878 & 7.8884 & 8.2593 & 8.1179 & 8.0391 \\
\cmidrule{2-10}
 & \multirow{2}{*}{\textbf{Avg}} & MAE & \textbf{5.3064} & 5.3579 & 5.5039 & 5.4570 & 5.5758 & 5.4324 & 5.4487 \\
 &  & RMSE & \textbf{7.1725} & 7.2317 & 7.3449 & 7.4323 & 7.4477 & 7.3178 & 7.2781 \\
\hline
\multirow{10}{*}{\textbf{ETTh1}} & \multirow{2}{*}{96$\to$96} & MAE & 0.3906 & 0.3916 & 0.4175 & 0.4055 & 0.4129 & 0.4112 & 0.3944 \\
 &  & MSE & 0.3854 & 0.3876 & 0.4033 & 0.3938 & 0.4108 & 0.4044 & 0.3894 \\
\cmidrule{2-10}
 & \multirow{2}{*}{96$\to$192} & MAE & 0.4267 & 0.4353 & 0.4771 & 0.4577 & 0.4509 & 0.4387 & 0.4399 \\
 &  & MSE & 0.4408 & 0.4584 & 0.5284 & 0.4790 & 0.4729 & 0.4644 & 0.4657 \\
\cmidrule{2-10}
 & \multirow{2}{*}{96$\to$336} & MAE & 0.4375 & 0.5054 & 0.4972 & 0.4499 & 0.5032 & 0.4875 & 0.4597 \\
 &  & MSE & 0.4551 & 0.5990 & 0.5508 & 0.5196 & 0.5703 & 0.5388 & 0.5169 \\
\cmidrule{2-10}
 & \multirow{2}{*}{96$\to$720} & MAE & 0.4904 & 0.6504 & 0.5993 & 0.5696 & 0.5792 & 0.5197 & 0.5166 \\
 &  & MSE & 0.5053 & 0.7988 & 0.6784 & 0.6405 & 0.6672 & 0.5654 & 0.5585 \\
\cmidrule{2-10}	
 & \multirow{2}{*}{\textbf{Avg}} & MAE & \textbf{0.4363} & 0.4957 & 0.4978 & 0.4707 & 0.4866 & 0.4643 & 0.4527 \\
 &  & MSE & \textbf{0.4467} & 0.5610 & 0.5402 & 0.5082 & 0.5303 & 0.4933 & 0.4826 \\
\hline
\multirow{10}{*}{\textbf{Electricity}} & \multirow{2}{*}{96$\to$96} & MAE & 0.2317 & 0.2319 & 0.2327 & 0.2456 & 0.2378 & 0.2424 & 0.2334 \\
 &  & MSE & 0.1397 & 0.1405 & 0.1404 & 0.1580 & 0.1464 & 0.1519 & 0.1415 \\
\cmidrule{2-10}
 & \multirow{2}{*}{96$\to$192} & MAE & 0.2434 & 0.2458 & 0.2480 & 0.2524 & 0.2590 & 0.2480 & 0.2443 \\
 &  & MSE & 0.1519 & 0.1551 & 0.1588 & 0.1634 & 0.1729 & 0.1574 & 0.1536 \\
\cmidrule{2-10}
 & \multirow{2}{*}{96$\to$336} & MAE & 0.2609 & 0.2613 & 0.2647 & 0.2724 & 0.2692 & 0.2665 & 0.2616 \\
 &  & MSE & 0.1685 & 0.1685 & 0.1717 & 0.1827 & 0.1809 & 0.1764 & 0.1689 \\
\cmidrule{2-10}
 & \multirow{2}{*}{96$\to$720} & MAE & 0.3004 & 0.3099 & 0.3014 & 0.3127 & 0.3014 & 0.3053 & 0.3141 \\
 &  & MSE & 0.2130 & 0.2185 & 0.2158 & 0.2363 & 0.2189 & 0.2244 & 0.2172 \\
\cmidrule{2-10}
 & \multirow{2}{*}{\textbf{Avg}} & MAE & \textbf{0.2591} & 0.2622 & 0.2617 & 0.2708 & 0.2669 & 0.2656 & 0.2634 \\
 &  & MSE & \textbf{0.1683} & 0.1707 & 0.1717 & 0.1851 & 0.1798 & 0.1775 & 0.1703 \\
\hline
\multirow{10}{*}{\textbf{Milan-SMS}} & \multirow{2}{*}{12$\to$12} & MAE & 13.5682 & 13.9090 & 14.3917 & 14.9624 & 14.9921 & 13.8966 & 13.9278 \\
 &  & RMSE & 32.3681 & 33.4033 & 32.7520 & 34.5235 & 35.1081 & 32.7695 & 32.9304 \\
\cmidrule{2-10}
 & \multirow{2}{*}{12$\to$24} & MAE & 13.1261 & 13.3298 & 13.2461 & 14.2761 & 13.1466 & 14.2761 & 13.4776 \\
 &  & RMSE & 32.4810 & 32.5360 & 32.6650 & 33.9048 & 32.6828 & 33.9048 & 32.7723 \\
\cmidrule{2-10}
 & \multirow{2}{*}{12$\to$48} & MAE & 14.8748 & 14.9251 & 14.8813 & 15.4177 & 15.5377 & 15.0831 & 14.8890 \\
 &  & RMSE & 35.6124 & 37.4446 & 37.0501 & 39.4167 & 39.7182 & 36.1761 & 36.2118 \\
\cmidrule{2-10}
 & \multirow{2}{*}{12$\to$96} & MAE & 17.7674 & 18.4134 & 17.7425 & 17.8328 & 18.4476 & 17.7855 & 17.8347 \\
 &  & RMSE & 39.5692 & 39.8092 & 41.2613 & 42.8343 & 46.5850 & 45.3674 & 41.7360 \\
\cmidrule{2-10}
 & \multirow{2}{*}{\textbf{Avg}} & MAE & \textbf{14.8341} & 15.1443 & 15.0654 & 15.6223 & 15.5310 & 15.2603 & 15.0323 \\
 &  & RMSE & \textbf{35.0077} & 35.7983 & 35.9321 & 37.6698 & 38.5235 & 37.0545 & 35.9126 \\
\hline
\multirow{10}{*}{\textbf{Milan-Call}} & \multirow{2}{*}{12$\to$12} & MAE & 9.3960 & 10.5166 & 9.7427 & 13.5081 & 9.6694 & 10.4797 & 10.4074 \\
 &  & RMSE & 24.2080 & 26.2404 & 24.8127 & 31.8630 & 25.0816 & 25.4622 & 25.9183 \\
\cmidrule{2-10}
 & \multirow{2}{*}{12$\to$24} & MAE & 9.6919 & 10.8491 & 10.1162 & 11.5436 & 10.1944 & 11.0501 & 10.5855 \\
 &  & RMSE & 26.7106 & 28.4898 & 27.1202 & 30.9409 & 27.1978 & 30.3064 & 28.7491 \\
\cmidrule{2-10}
 & \multirow{2}{*}{12$\to$48} & MAE & 12.7806 & 14.2654 & 13.2892 & 13.5407 & 13.6035 & 13.6961 & 13.1066 \\
 &  & RMSE & 33.3597 & 38.8279 & 36.7287 & 38.3853 & 35.9779 & 37.2930 & 35.0805 \\
\cmidrule{2-10}
 & \multirow{2}{*}{12$\to$96} & MAE & 13.7592 & 14.9443 & 15.6378 & 14.5347 & 16.3659 & 15.4093 & 15.7339 \\
 &  & RMSE & 34.9308 & 36.1698 & 44.6959 & 37.6496 & 37.3060 & 37.2042 & 36.2168 \\
\cmidrule{2-10}
 & \multirow{2}{*}{\textbf{Avg}} & MAE & \textbf{11.4069} & 12.6439 & 12.1965 & 13.2818 & 12.4583 & 12.6588 & 12.4584 \\
 &  & RMSE & \textbf{29.8023} & 32.4320 & 33.3394 & 34.7097 & 31.3908 & 32.5665 & 31.4912 \\
\hline
\multirow{10}{*}{\textbf{Milan-Internet}} & \multirow{2}{*}{12$\to$12} & MAE & 51.3435 & 62.9649 & 62.1237 & 98.3414 & 62.4344 & 67.7585 & 64.2594 \\
 &  & RMSE & 127.3646 & 147.3886 & 146.8808 & 220.6383 & 153.1805 & 157.0537 & 155.7520 \\
\cmidrule{2-10}
 & \multirow{2}{*}{12$\to$24} & MAE & 53.8845 & 66.7705 & 58.8052 & 61.2277 & 59.1715 & 68.7719 & 61.6381 \\
 &  & RMSE & 139.1029 & 168.3586 & 140.2052 & 154.3039 & 153.0713 & 165.2431 & 151.8447 \\
\cmidrule{2-10}
 & \multirow{2}{*}{12$\to$48} & MAE & 55.3451 & 74.8081 & 64.9854 & 76.2779 & 69.0151 & 70.2102 & 67.4565 \\
 &  & RMSE & 143.7303 & 176.8681 & 163.1520 & 186.1613 & 175.2776 & 170.2111 & 170.1674 \\
\cmidrule{2-10}
 & \multirow{2}{*}{12$\to$96} & MAE & 66.8640 & 80.2353 & 80.4594 & 78.9032 & 80.8676 & 80.4003 & 81.7781 \\
 &  & RMSE & 168.5287 & 188.3620 & 191.4470 & 190.6136 & 192.3180 & 187.3065 & 200.8724 \\
\cmidrule{2-10}
 & \multirow{2}{*}{\textbf{Avg}} & MAE & \textbf{56.8593} & 71.1947 & 66.5934 & 78.6876 & 67.8722 & 71.7852 & 68.7830 \\
 &  & RMSE & \textbf{144.6816} & 170.2443 & 160.4213 & 187.9293 & 168.4619 & 169.9536 & 169.6591 \\
\bottomrule
\end{tabular}
}
\end{table*}

\section{Limitations and future directions}\label{appendix:limit_future}
The datasets used by this work do not cover all spatio-temporal datasets. We will explore using larger dataset collections to further improve STM3.

STM3 gives a new idea for spatio-temporal modeling, particularly in scenarios requiring fine-grained, long-term predictions. We plan to build a foundation model using multiple spatio-temporal datasets using STM3's framework to further explore its scalability.

\end{document}